\documentclass{article}
\usepackage[final]{colm2026_conference}

\usepackage{microtype}
\usepackage{hyperref}
\usepackage{url}
\usepackage{booktabs}
\usepackage{multirow}
\usepackage{amsmath,amssymb,amsfonts}
\usepackage{enumitem}
\usepackage{colortbl}
\usepackage{pgfplots}
\pgfplotsset{compat=1.17}
\pgfplotsset{compat=1.18}
\usepgfplotslibrary{groupplots}
\usepackage{longtable}
\usepackage{pifont}
\usepackage{placeins}
\usepackage{float}
\usepackage{lineno}
\usetikzlibrary{arrows.meta, calc}

\definecolor{darkblue}{rgb}{0, 0, 0.5}
\hypersetup{colorlinks=true, citecolor=darkblue, linkcolor=darkblue, urlcolor=darkblue}

\definecolor{lightred}{HTML}{FFE0E0}
\definecolor{lightgreen}{HTML}{E0FFE0}
\title {In-Context Examples Suppress Scientific Knowledge Recall in LLMs}

\author{Chaemin Jang$^{1}$, Woojin Park$^{2}$, Hyeok Yun$^{1}$, Dongman Lee$^{1\dagger}$, Jihee Kim$^{1}$\thanks{Co-corresponding authors.} \\
$^{1}$KAIST, South Korea \quad $^{2}$Shanghai Jiao Tong University, China \\
\texttt{jchaemin@kaist.ac.kr}
}

\begin{document}

\ifcolmsubmission
\linenumbers
\fi

\setcounter{footnote}{1}
\maketitle

\begin{abstract}
Scientific reasoning rarely stops at what is directly observable; it often requires uncovering hidden structure from data. From estimating reaction constants in chemistry to inferring demand elasticities in economics, this \emph{latent structure recovery} is what distinguishes scientific reasoning from curve fitting.
Large language models (LLMs) can often recall and apply relevant scientific formulas, but we show that this ability is surprisingly easy to suppress.
We show that adding in-context examples makes models rely less on pretrained domain knowledge, even when those examples are generated by the very same formula. Rather than reinforcing knowledge-driven derivation, examples shift computation toward empirical pattern fitting. We document this \emph{knowledge displacement} on 60 latent structure recovery tasks across five scientific domains, 6,000 trials, and four models. 
This displacement is consistent across domains, but its accuracy consequences depend on how the displaced strategy compares to the one that replaces it: the same shift can lower accuracy, leave it unchanged, or appear to improve it. 
In all cases, however, the model shifts away from knowledge-driven reasoning. For practitioners deploying LLMs on scientific tasks, the message is cautionary: in-context examples may displace, rather than reinforce, the knowledge they are intended to support.
\end{abstract}

\section{Introduction}
\label{sec:intro}
 
Since \citet{brown2020language} showed that in-context examples can substantially improve model performance, and \citet{wei2022chain} demonstrated that chain-of-thought examples can elicit multi-step reasoning, in-context learning (ICL) has become a standard component of scientific LLM systems. It is now used in settings ranging from multi-agent pipelines for materials science \citep{ghafarollahi2024sciagents} to equation discovery from data \citep{shojaee2025llmsr,shojaee2024llmsrbench} to scientific problem-solving benchmarks spanning biology, chemistry, and physics \citep{du2025sde,luo2025llm4sr}. Despite their differences, these systems share two design choices: they provide in-context examples by default, and they evaluate performance primarily through end-to-end accuracy. Yet it remains unclear whether in-context examples help models retrieve and apply pretrained domain knowledge on scientific tasks that require it.

Prior work suggests that in-context examples do not simply reinforce what a model already knows. \citet{min2022rethinking} showed that replacing correct labels with random ones leaves ICL performance largely intact, suggesting that demonstrations often serve as task cues rather than as sources of new knowledge. \citet{wei2023larger} found that larger models can override their own semantic priors when contextual labels conflict with stored knowledge. Both studies, however, examine settings in which context diverges from, or competes with, what the model already knows. We study a different and previously untested condition: examples that are fully consistent with a model’s pretrained scientific knowledge, yet suppress its use rather than reinforce it.

\begin{figure}[t]
\centering
\resizebox{\textwidth}{!}{%
\begin{tikzpicture}[
  font=\footnotesize,
  pbox/.style={line width=0.6pt, rounded corners=3pt,
               align=center, minimum height=2.2cm, text width=2.2cm, inner sep=5pt},
  mbox/.style={line width=0.6pt, rounded corners=3pt,
               align=center, minimum height=2.2cm, text width=2.5cm, inner sep=5pt},
  inbox/.style ={pbox, fill=gray!10,   draw=gray!75},
  midbox/.style={mbox, fill=cknow!18,  draw=cknow!75},
  outbox/.style={pbox, fill=gray!10,   draw=gray!75},
  harr/.style={-{Latex[length=5pt,width=5pt]}, line width=0.9pt, black},
  head/.style={font=\large\bfseries, text=black, align=center},
  sub/.style={font=\footnotesize, align=center},
  under/.style={font=\footnotesize, align=center},
]
\definecolor{cknow}{HTML}{2C4A6E}

\node[head] at (3.22,2.95) {Knowledge-Driven Reasoning ($\mathcal{S}_{\text{know}}$)};
\node[sub]  at (3.22,1.95)
  {Recognize task, recall domain law.\\
   Apply Newton's law:\\
   $T(t)=T_{\text{env}}+(T_0-T_{\text{env}})e^{-kt}$};
\node[inbox]  (LA) at (0,0)
  {$[90,72,59]\Rightarrow\,?$};
\node[midbox] (LB) at (3.22,0)
  {Room temp\\ $T_{\text{env}}=25.2^\circ$C\\[4pt]
   Cooling rate\\ $k=0.0651$};
\node[outbox] (LC) at (6.44,0)
  {How long until\\ $T=30^\circ$C?\\[6pt]
   $t=\textbf{40.0 min}$~\textcolor{green!45!black}{\checkmark}};
\draw[harr] (LA) -- (LB);
\draw[harr] (LB) -- (LC);
\node[under] at (0,-1.40)    {$X_{\text{test}}$};
\node[under] at (3.22,-1.40) {Latents $\Theta$};
\node[under] at (6.44,-1.40) {Target $Y$};

\draw[black!50, line width=0.5pt] (8.33,-1.65) -- (8.33,3.30);

\node[head] at (13.44,2.95) {ICL-Driven Pattern Fitting ($\mathcal{S}_{\text{emp}}$)};
\node[sub]  at (13.44,1.95)
  {Analyze numerical trends.\\
   Bypasses domain law.\\
   Directly approximate $Y$ from the test input.};
\node[inbox]  (RA) at (10.22,0)
  {{\scriptsize $[86,70,58]\to 36.1$}\\[1pt]
   {\scriptsize $[95,71,55]\to 28.7$}\\[1pt]
   {\scriptsize $[80,62,50]\to 32.1$}\\[1pt]
   {\scriptsize\makebox[2.0cm]{\dotfill}}\\[1pt]
   $[90,72,59]\Rightarrow\,?$};
\node[midbox] (RB) at (13.44,0)
  {Fit from\\ demonstrations:\\[4pt]
   $\hat{t}=f(T_0,T_5,T_{10})$};
\node[outbox] (RC) at (16.66,0)
  {How long until\\ $T=30^\circ$C?\\[6pt]
   Estimated $t$\\
   $\approx\textbf{32.8 min}$~\textcolor{red!60!black}{$\times$}};
\draw[harr] (RA) -- (RB);
\draw[harr] (RB) -- (RC);
\node[under] at (10.22,-1.40) {$X_{\text{train}}$};
\node[under] at (13.44,-1.40) {Empirical Pattern};
\node[under] at (16.66,-1.40) {Target $Y$};

\end{tikzpicture}%
}
\caption{Two competing reasoning modes on Newton's cooling. \textbf{Left} ($\mathcal{S}_{\text{know}}$): the model recognizes
the domain, recalls Newton's law, extracts latent parameters from the test
observables, and solves for the target time analytically. \textbf{Right}
($\mathcal{S}_{\text{emp}}$): given in-context examples, the model bypasses the domain law and instead fits an empirical trend to approximate the target directly.}
\label{fig:reasoning_modes}
\end{figure}

We investigate this phenomenon on tasks requiring \emph{latent structure recovery}. In these tasks, the model receives raw observables, must infer hidden parameters from them using domain knowledge, and then apply a known formula to compute a target quantity. For example, given temperature readings from a cooling object, the model must recover the ambient temperature and cooling rate, then use Newton’s cooling law to predict when the object will reach a specified temperature (Figure~\ref{fig:reasoning_modes}). This observe-extract-compute structure is common in quantitative scientific reasoning: extracting reaction constants from kinetic data, recovering elastic moduli from stress-strain relationships, or estimating demand elasticities from market observations. In each case, success depends not merely on mapping inputs to outputs, but on recalling and applying the scientific law that governs the domain.

LLMs can often recall and apply these formulas from pretrained memory. We show that in-context examples suppress this scientific knowledge recall, even when those examples are generated by the very formula the model already knows. To explain this, we propose a strategy competition framework in which two computational modes operate on these tasks: knowledge-driven derivation, which retrieves a domain formula from pretrained memory and derives the answer through it, and example-driven fitting, which approximates the input-output mapping from the provided examples without invoking domain knowledge.

Across 60 latent structure recovery tasks spanning five scientific domains, 6,000 trials, and four models, adding a few in-context examples consistently shifts computation away from knowledge-driven derivation and toward example-driven fitting --- a shift we refer to as \emph{knowledge displacement}. The accuracy consequences vary across domains --- declining in some, appearing to improve in others, and remaining flat elsewhere --- but the underlying displacement is direction-consistent, and aggregate accuracy alone obscures it (\S\ref{sec:mechanism}). Examples that are fully consistent with a model's pretrained scientific knowledge can therefore displace, rather than reinforce, knowledge-driven derivation.

\paragraph{Contributions.}
\begin{enumerate}[leftmargin=*,itemsep=1pt]
\item We show that in-context examples consistently shift scientific latent structure recovery away from knowledge-driven derivation and toward example-driven fitting, even when the examples are fully consistent with the underlying domain formula  (\S\ref{sec:results}).
\item We introduce a strategy competition analysis that decomposes the apparently contradictory accuracy effects across domains into three regimes determined by the quality gap between the displaced and replacement strategies (\S\ref{sec:mechanism}).
\item We construct a difficulty-balanced benchmark of 60 latent structure recovery tasks across five scientific domains, designed to make this shift empirically observable (\S\ref{sec:setup}).
\end{enumerate}

\section{Related work}
\label{sec:related}

\textbf{What ICL computes.}
Theoretical work characterizes ICL as implicit Bayesian inference over a hypothesis space shaped by pretraining \citep{xie2022explanation}, or as gradient descent executed within the forward pass \citep{garg2022can,dai2023gpt,vonoswald2023transformers}. Mechanistic studies identify induction heads \citep{olsson2022context,yin2025attention}, and show that demonstrations causally drive ICL behavior \citep{hendel2023task,todd2024function, akyurek2022what,li2023transformers}. A key assumption across this literature is that the target function lies within the model's hypothesis space. Our benchmark violates this: composite functions $h(g(X))$ require multi-step latent isolation that cannot be solved by regression over examples, as confirmed by a boundary test (Appendix~\ref{app:boundary}).

\textbf{Task recognition versus task learning.}
\citet{pan2023context} decomposed ICL into \emph{task recognition} (applying pretrained knowledge) and \emph{task learning} (inferring new mappings from demonstrations). \citet{min2022rethinking} showed that replacing gold labels with random labels barely hurts ICL performance, suggesting models use demonstrations primarily for task recognition. \citet{kossen2024incontext} qualify this, showing that ICL does learn label relationships in settings that isolate them, so the random-label result bounds rather than eliminates task learning. Our ``strategy competition'' framework refines this decomposition for knowledge recall in scientific tasks: knowledge-driven derivation corresponds to task recognition (the model identifies ``this is a Cobb-Douglas problem'' and recalls the formula), while example-driven fitting corresponds to task learning (the model fits patterns from examples). The novel contribution is showing that these two modes \emph{compete}, and that examples suppress the former in favor of the latter.

\textbf{When priors and context conflict.}
\citet{wei2023larger} showed that larger models override semantic priors more effectively when labels are flipped, establishing that the prior-context balance is scale-dependent. \citet{chan2022data} demonstrated that data distributional properties determine which ICL behaviors emerge during pretraining. \citet{razeghi2022impact} found that pretraining term frequencies predict few-shot numerical reasoning accuracy. More recently, \citet{tang2023large} showed that LLMs exploit shortcuts in ICL, relying on spurious correlations rather than learning the intended task structure. These works establish that priors matter for ICL, but all study settings in which context and priors conflict \citep{longpre2021entity, xie2024adaptive}. We show that displacement occurs even when they do not: in-context examples suppress knowledge-driven derivation even when the examples are fully consistent with the model's stored formula.

\textbf{Scientific reasoning benchmarks.}
Existing benchmarks evaluate scientific knowledge recall \citep{hendrycks2021math,rein2024gpqa,wang2024scibench}, counterfactual robustness \citep{wu2024reasoning}, economic reasoning \citep{raman2024steer,econnli2024}, and multi-step scientific discovery \citep{du2025sde}. Our benchmark differs in requiring explicit latent variable isolation ($X \to \Theta \to Y$): the model cannot reach the correct answer by surface pattern matching or formula application alone; it must first recover hidden parameters from observables, then compute through them. This structure makes strategy competition visible in the chain-of-thought trace.

\section{Benchmark and experimental setup}
\label{sec:setup}

\subsection{Problem formulation and hypothesis}

Standard ICL evaluations assess a direct mapping $f: X \to Y$. We formalize a harder problem: \textbf{Latent Structure Recovery}. Given observables $X$, the model must induce $g: X \to \Theta$ (latent extraction) then apply $h: \Theta \to Y$ (target computation), so $f(X) = h(g(X))$.
 
We hypothesize a \emph{Strategy Competition}: when the model faces a latent structure recovery task, two strategies compete. \textbf{Knowledge-driven derivation} ($\mathcal{S}_{\text{know}}$) retrieves domain-specific equations from pretrained memory, triggered by semantic cues, and derives the answer through the recalled formula. \textbf{Example-driven fitting} ($\mathcal{S}_{\text{emp}}$) bypasses domain knowledge and instead approximates $f(X)$ via pattern-fitting over $k$-shot examples. Our core claim is that adding examples does not combine these strategies; it shifts the balance from $\mathcal{S}_{\text{know}}$ toward $\mathcal{S}_{\text{emp}}$ across domains.

\begin{figure}[t]
\centering
\hspace*{-0.5cm}
\begin{tikzpicture}[
  font=\small,
  box/.style={draw=gray!75, line width=0.6pt, rounded corners=1.5pt,
              fill=gray!10, minimum width=1.4cm, minimum height=0.62cm, align=center},
  lbox/.style={box, fill=cknow!18, draw=cknow!75},
  arr/.style={-{Latex[length=4pt,width=4pt]}, line width=0.7pt, black!80},
  note/.style={font=\footnotesize, black, align=center},
  rowlab/.style={font=\footnotesize, black, align=center},
  fnote/.style={font=\footnotesize, black!80, align=center},
]
\definecolor{cknow}{HTML}{2C4A6E}
\node[rowlab] at (1.4,1.25) {Standard ICL};
\node[rowlab] at (1.4,0)    {Latent recovery\\(ours)};
\node[box, fill=gray!5, minimum height=0.55cm] (Xs) at (4.0,1.25) {$X$};
\node[box, fill=gray!5, minimum height=0.55cm] (Ys) at (11.8,1.25) {$Y$};
\draw[arr, black!60] (Xs) -- node[above=1pt, note]{$f$: direct mapping} (Ys);
\node[box]  (X)  at (4.0,0)  {$X$};
\node[lbox] (Th) at (7.9,0)  {$\Theta$};
\node[box]  (Y)  at (11.8,0) {$Y$};
\node[fnote] (xn) at (4.0,-0.80)  {observables\\(3--8 per task)};
\node[fnote] (tn) at (7.9,-0.80)  {latent parameters\\(1--4 per task)};
\node[fnote] (yn) at (11.8,-0.80) {target};
\draw[arr] (X)  -- node[above=1pt, note]{$g$: extraction\\(8 forms)}   (Th);
\draw[arr] (Th) -- node[above=1pt, note]{$h$: computation\\(12 forms)} (Y);
\draw[line width=0.6pt, black!60, dashed, dash pattern=on 1.8pt off 1.8pt,
      -{Latex[length=4pt,width=4pt]}]
  (xn.south) .. controls (6.1,-2.05) and (9.7,-2.05) .. (yn.south)
  node[pos=0.5, fill=white, inner sep=1.5pt, font=\large]
    {\textcolor{red!60!black}{$\times$}};
\node[note] at (7.9,-2.35)
  {not solvable by regression over demonstrations (Appendix~\ref{app:boundary})};
\end{tikzpicture}
\caption{Task structure. Our benchmark requires recovering latent parameters before computing the target, $f(X)=h(g(X))$. The
composite is not reachable by pattern fitting over demonstrations
(boundary test, Appendix~\ref{app:boundary}). Tasks span 8
observable-to-latent and 12 latent-to-target equation forms with 1--4
latents per task (Appendix~\ref{app:full_inventory}).}
\label{fig:task_structure}
\end{figure}

\subsection{Benchmark and evaluation}

\subsubsection{Task design}

We constructed a 60-task benchmark spanning Biology, Chemistry, Economics, Physics, and Geoscience (12 per domain). The resulting zero-shot accuracies range from 55\% to 88\%, providing a natural spectrum of prior quality that allows us to observe how displacement interacts with the strength of the model's existing knowledge. Each task enforces a latent-observable architecture $(X, \Theta, Y, g, h)$: the model receives observables $X$, must extract $\Theta = g(X)$ via a domain law, then compute $Y = h(\Theta)$. Tasks implement canonical equations from introductory or intermediate university courses (e.g., Nernst equation, Cobb-Douglas production, Snell's law, Darcy's law). The full task inventory is in Appendix~\ref{app:full_inventory}.

Tasks are characterized by four structural complexity criteria: number of observables, number of latent variables, observable-to-latent equation format, and latent-to-target format. These span 8 observable-to-latent equation formats, 12 latent-to-target formats, and 1--4 latent variables per task, ensuring that our findings reflect diverse computational structures rather than artifacts of a particular equation type. The benchmark is difficulty-balanced across domains (Kruskal-Wallis $p = 0.44$; tier distribution in Appendix~\ref{app:tasks}).

\subsubsection{Data generation, evaluation conditions, and strategy classifier}

For each task, we generate problem instances by sampling latent parameters $\Theta$ uniformly from physically plausible ranges. Given $\Theta$, we compute observables $X$ by evaluating the preimage of $g$, $g^{-1}(\Theta)$, and the ground-truth target $Y = h(\Theta)$ analytically. Each of the 50 test trials per task uses independently sampled parameters, ensuring that the model cannot memorize specific numerical values. In the 10-shot condition, the 10 training examples are generated by the same procedure --- same formula, independently sampled parameters --- so the examples are fully consistent with the domain law the model already knows.
%

All primary experiments used GPT-5.2 under Zero-Shot CoT \citep{kojima2022large} and 10-Shot CoT, with 50 independently generated trials per task (6,000 total). The 10-shot prompt provides input-output pairs without intermediate derivations, reflecting standard deployment practice where practitioners supply observational data rather than worked solutions. Demonstrations that contain the derivation itself primarily test procedural execution rather than knowledge retrieval; a worked-demonstration control appears in Appendix~\ref{app:worked_control}. Full prompt templates for every condition appear in Appendix~\ref{app:prompts}, model versions and decoding parameters in Appendix~\ref{app:implementation}.
%

GPT-4o classified each CoT trajectory as \textsc{Parametric}, \textsc{Empirical}, or \textsc{Hybrid} (Appendix~\ref{app:validation}). Two human annotators established ground-truth labels on 100 traces; the automated classifier achieved 96\% agreement with human consensus. Appendix~\ref{app:two_axis} additionally re-classifies trajectories along two binary axes, correct-law and correct-numerics.

\section{Results: knowledge displacement across domains}
\label{sec:results}

We first report the domain-level accuracy effects of adding 10 in-context examples.

\begin{table}[t]
\centering
\begin{tabular}{lcccc}
\toprule
\textbf{Domain} & \textbf{Zero-Shot} & \textbf{10-Shot} & $\boldsymbol{\Delta}$\textbf{(pp)} & $\boldsymbol{p}$ \\
\midrule
Economics  & 88.3\% & 72.7\% & \cellcolor{lightred}\textbf{$-$15.7} & $<$.001 \\
Chemistry  & 66.0\% & 59.8\% & \cellcolor{lightred}\textbf{$-$6.2} & 0.007 \\
Physics    & 85.0\% & 85.2\% & $+$0.2 & 1.00 \\
Biology    & 71.2\% & 72.5\% & $+$1.3 & 0.54 \\
Geoscience & 55.3\% & 68.7\% & \cellcolor{lightgreen}\textbf{+13.3} & $<$.001 \\
\midrule
\textbf{Overall} & 73.2\% & 71.8\% & $-$1.4 & \\
\bottomrule
\end{tabular}
\caption{Domain-level accuracy (50 trials $\times$ 12 tasks per domain). $p$-values from McNemar's test. Per-task results for all 60 tasks are in Appendix~\ref{app:per_task}.}
\label{tab:domain}
\end{table}

\subsection{The domain-wise inversion}

Table~\ref{tab:domain} shows the central result. Adding 10 in-context examples degrades accuracy in Economics ($-15.7$pp, $p < .001$) and Chemistry ($-6.2$pp, $p = .007$), while Geoscience shows apparent improvement ($+13.3$pp, $p < .001$). Biology and Physics show no significant change, and the aggregate ($-$1.4pp) masks the divergence entirely. We explain why the same underlying shift produces these opposing accuracy effects in \S\ref{sec:mechanism}. The pattern holds under a random-effects meta-analysis accounting for task-level variability (Appendix~\ref{app:stats}), within fraction-format tasks alone (Appendix~\ref{app:confound}), after excluding ceiling tasks (Appendix~\ref{app:ceiling_analysis}), and after dropping the two most extreme tasks per domain (Appendix~\ref{app:sensitivity}).

\subsection{Cross-model replication}

We replicated on Claude Sonnet~4.6, Llama~4~Maverick, and DeepSeek~V3 (30 trials per task). Economics is harmed in all four models (4/4 negative); Geoscience shows apparent gains consistent with displacement in all four (4/4 positive). Middle domains show small, noisy effects (Figure~\ref{fig:main_results}; full numerical results in Appendix~\ref{app:cross_model}).

\begin{figure*}[t]
\centering
\begin{tikzpicture}
\begin{groupplot}[
    group style={
        group size=2 by 2,
        horizontal sep=1.4cm,
        vertical sep=1.4cm,
        ylabels at=edge left,
    },
    ybar,
    width=0.50\textwidth,
    height=4.2cm,
    ylabel={Accuracy (\%)},
    ylabel style={font=\small},
    symbolic x coords={Econ, Chem, Phys, Bio, Geo},
    xtick=data,
    xticklabel style={font=\small, rotate=35, anchor=east},
    yticklabel style={font=\small},
    ymin=0, ymax=115,
    ytick={0,20,40,60,80,100},
    enlarge x limits=0.18,
    grid=none,
    axis line style={gray!50},
    every axis plot/.append style={fill opacity=0.85},
    legend style={at={(0.5,1.05)}, anchor=south, font=\small, draw=none, legend columns=2},
    title style={font=\small\bfseries, yshift=-2pt},
]
 
\nextgroupplot[title={GPT-5.2}]
\addplot[fill=blue!50, draw=blue!70!black] coordinates {
    (Econ, 88.3) (Chem, 66.0) (Phys, 85.0) (Bio, 71.2) (Geo, 55.3)};
\addplot[fill=orange!50, draw=orange!70!black] coordinates {
    (Econ, 72.7) (Chem, 59.8) (Phys, 85.2) (Bio, 72.5) (Geo, 68.7)};
\node[font=\small\bfseries, red!70!black] at (axis cs:Econ,97) {$-$15.7};
\node[font=\small\bfseries, red!70!black] at (axis cs:Chem,74) {$-$6.2};
\node[font=\small, gray] at (axis cs:Phys,93) {$+$0.2};
\node[font=\small, gray] at (axis cs:Bio,80) {$+$1.3};
\node[font=\small\bfseries, green!50!black] at (axis cs:Geo,76) {$+$13.3};
 
\nextgroupplot[title={Claude 4.6}]
\addplot[fill=blue!50, draw=blue!70!black] coordinates {
    (Econ, 100.0) (Chem, 94.7) (Phys, 94.7) (Bio, 98.3) (Geo, 75.0)};
\addplot[fill=orange!50, draw=orange!70!black] coordinates {
    (Econ, 93.1) (Chem, 90.9) (Phys, 95.6) (Bio, 98.9) (Geo, 90.3)};
\node[font=\small\bfseries, red!70!black] at (axis cs:Econ,108) {$-$6.9};
\node[font=\small\bfseries, red!70!black] at (axis cs:Chem,103) {$-$3.9};
\node[font=\small, gray] at (axis cs:Phys,104) {$+$0.9};
\node[font=\small, gray] at (axis cs:Bio,107) {$+$0.6};
\node[font=\small\bfseries, green!50!black] at (axis cs:Geo,98) {$+$15.3};
 
\nextgroupplot[title={Llama 4}]
\addplot[fill=blue!50, draw=blue!70!black] coordinates {
    (Econ, 81.4) (Chem, 87.8) (Phys, 91.9) (Bio, 72.8) (Geo, 63.1)};
\addplot[fill=orange!50, draw=orange!70!black] coordinates {
    (Econ, 71.1) (Chem, 93.1) (Phys, 89.4) (Bio, 74.7) (Geo, 81.4)};
\node[font=\small\bfseries, red!70!black] at (axis cs:Econ,89) {$-$10.3};
\node[font=\small\bfseries, green!50!black] at (axis cs:Chem,101) {$+$5.3};
\node[font=\small, gray] at (axis cs:Phys,100) {$-$2.5};
\node[font=\small, gray] at (axis cs:Bio,82) {$+$1.9};
\node[font=\small\bfseries, green!50!black] at (axis cs:Geo,89) {$+$18.3};
 
\nextgroupplot[title={DeepSeek V3}]
\addplot[fill=blue!50, draw=blue!70!black] coordinates {
    (Econ, 78.1) (Chem, 78.1) (Phys, 91.1) (Bio, 78.9) (Geo, 62.5)};
\addplot[fill=orange!50, draw=orange!70!black] coordinates {
    (Econ, 71.7) (Chem, 76.9) (Phys, 89.7) (Bio, 80.5) (Geo, 72.4)};
\node[font=\small\bfseries, red!70!black] at (axis cs:Econ,86) {$-$6.4};
\node[font=\small, gray] at (axis cs:Chem,86) {$-$1.2};
\node[font=\small, gray] at (axis cs:Phys,99) {$-$1.4};
\node[font=\small, gray] at (axis cs:Bio,88) {$+$1.7};
\node[font=\small\bfseries, green!50!black] at (axis cs:Geo,80) {$+$10.0};
 
\end{groupplot}
\node[anchor=south, font=\small] at ($(group c1r1.north)!0.5!(group c2r1.north) + (0, 0.6)$) {%
    \tikz{\fill[blue!50, draw=blue!70!black, opacity=0.85] (0,0) rectangle (0.25,0.2);} Zero-shot \quad
    \tikz{\fill[orange!50, draw=orange!70!black, opacity=0.85] (0,0) rectangle (0.25,0.2);} 10-shot};
\end{tikzpicture}
\caption{Cross-model replication (30 trials per task, except GPT-5.2: 50 trials). Blue = zero-shot, orange = 10-shot. Numbers above bars show $\Delta$ (pp). Economics is negative across all four models; Geoscience is positive across all four.}
\label{fig:main_results}
\end{figure*}

\section{Analysis: strategy competition explains why displacement 
is direction-consistent but its consequences are not}
\label{sec:mechanism}

We now show that strategy displacement from $\mathcal{S}_{\text{know}}$ 
toward $\mathcal{S}_{\text{emp}}$ explains all five domain-level outcomes 
in \S\ref{sec:results}. We first establish that displacement is consistent in direction 
across domains regardless of its accuracy consequence 
(\S\ref{subsec:strategy_shift}), then explain why the same displacement 
produces different accuracy effects depending on the quality of the 
displaced strategy (\S\ref{subsec:accuracy_effects}), and finally probe 
why $\mathcal{S}_{\text{know}}$ is easily suppressed by the mere 
presence of examples (\S\ref{subsec:decoy}).

\subsection{ICL triggers a direction-consistent strategy shift}
\label{subsec:strategy_shift}

\begin{table}[t]
\centering
\setlength{\tabcolsep}{3pt}
{\small
\begin{tabular}{llcccc}
\toprule
\textbf{Domain} & \textbf{Cond.} & \textbf{Param.} & \textbf{Empir.} & \textbf{Hybrid} & $\boldsymbol{\Delta}$\textbf{Acc.} \\
\midrule
\multirow{2}{*}{Economics}  & 0-shot  & 100.0\% & 0.0\% & 0.0\%  & \multirow{2}{*}{\textcolor{red!70!black}{\textbf{$-$15.7}}} \\
                            & 10-shot & 80.3\%  & 2.0\% & 17.7\% & \\
\midrule
\multirow{2}{*}{Chemistry}  & 0-shot  & 91.8\%  & 7.7\% & 0.5\% & \multirow{2}{*}{\textcolor{red!70!black}{\textbf{$-$6.2}}} \\
                            & 10-shot & 81.3\%  & 13.0\% & 5.7\% & \\
\midrule
\multirow{2}{*}{Physics}    & 0-shot  & 99.8\% & 0.0\% & 0.2\% & \multirow{2}{*}{+0.2} \\
                            & 10-shot & 93.8\% & 2.2\% & 4.0\% & \\
\midrule
\multirow{2}{*}{Biology}    & 0-shot  & 87.8\% & 10.7\% & 1.5\% & \multirow{2}{*}{$+$1.3} \\
                            & 10-shot & 74.7\% & 14.3\% & 11.0\% & \\
\midrule
\multirow{2}{*}{Geoscience} & 0-shot  & 85.7\%  & 13.2\% & 1.2\%  & \multirow{2}{*}{\textcolor{green!50!black}{\textbf{+13.3}}} \\
                            & 10-shot & 65.7\%  & 19.0\% & 15.3\%  & \\
\bottomrule
\end{tabular}
}
\caption{Strategy distribution from automated CoT classification (\S\ref{sec:setup}). Columns show the share of trials classified as each strategy. ICL reduces $\mathcal{S}_{\text{know}}$ (Parametric) by 6--20pp in every domain, regardless of the accuracy consequence (rightmost column).}
\label{tab:strategy}
\end{table}

Table~\ref{tab:strategy} tests our core prediction directly. In every domain, adding examples reduces the share of $\mathcal{S}_{\text{know}}$ trajectories by 6--20pp, shifting trials toward $\mathcal{S}_{\text{emp}}$ or hybrid strategies. Critically, this shift is not adaptive: the model 
does not selectively abandon knowledge recall in domains where it would be wrong and preserve it where it would be right. Economics, where 
$\mathcal{S}_{\text{know}}$ achieves 88.3\% accuracy, loses 19.7pp of 
parametric trajectories; Geoscience, where $\mathcal{S}_{\text{know}}$ 
achieves only 54.1\%, loses 20.0pp. The magnitude of displacement bears 
no relationship to the quality of the displaced strategy. The accuracy 
consequence differs across domains, but the displacement mechanism 
does not. It is also not specific to $k = 10$: 
sweeping $k \in \{0, 1, 5, 10, 20\}$ (Table~\ref{tab:ksweep}) shows that 
each domain's effect direction emerges and saturates by 
$k = 10$, with $k = 20$ within roughly 3pp of $k = 10$ in every domain, 
so displacement is a few-shot phenomenon rather than an artifact of one 
shot count. A paired logit-difference analysis shows the same shift below the level of text: demonstrations move probability mass toward the example-fit answer and away from the knowledge answer (Table~\ref{tab:logit_diff}, Appendix~\ref{app:representational}).

\begin{table}[t]
\centering
{\small
\begin{tabular}{lccccc}
\toprule
\textbf{Domain} & $\boldsymbol{k=0}$ & $\boldsymbol{k=1}$ & $\boldsymbol{k=5}$ & $\boldsymbol{k=10}$ & $\boldsymbol{k=20}$ \\
\midrule
Economics  & 88.3 & 76.4 & 74.7 & 72.7 & 74.2 \\
Chemistry  & 66.0 & 61.4 & 63.9 & 59.8 & 61.1 \\
Physics    & 85.0 & 83.6 & 86.1 & 85.2 & 83.6 \\
Biology    & 71.2 & 73.3 & 70.8 & 72.5 & 71.7 \\
Geoscience & 55.3 & 65.8 & 63.1 & 68.7 & 66.4 \\
\bottomrule
\end{tabular}
}
\caption{Accuracy (\%) across $k$ (GPT-5.2; 30 trials/task/cell; $k=0$ and $k=10$ from the main experiment).}
\label{tab:ksweep}
\end{table}

\subsection{Why knowledge displacement produces different accuracy effects}
\label{subsec:accuracy_effects}

\begin{table}[t]
\centering
\setlength{\tabcolsep}{3pt}
{\small
\begin{tabular}{llrrrrrr}
\toprule
& & \multicolumn{2}{c}{\textbf{Parametric}} & \multicolumn{2}{c}{\textbf{Empirical}} & \multicolumn{2}{c}{\textbf{Hybrid}} \\
\cmidrule(lr){3-4} \cmidrule(lr){5-6} \cmidrule(lr){7-8}
\textbf{Domain} & \textbf{Cond.} & \textbf{Acc.} & $n$ & \textbf{Acc.} & $n$ & \textbf{Acc.} & $n$ \\
\midrule
\multirow{2}{*}{Econ.}  & 0-shot  & \textbf{88.3} & 600 & --- & 0  & --- & 0 \\
                        & 10-shot & \textbf{78.2} & 482 & 41.7 & 12 & 50.9 & 106 \\
\midrule
\multirow{2}{*}{Chem.}  & 0-shot  & \textbf{70.6} & 551 & 13.0 & 46 & 33.3 & 3 \\
                         & 10-shot & \textbf{65.2} & 488 & 33.3 & 78 & 44.1 & 34 \\
\midrule
\multirow{2}{*}{Phys.}  & 0-shot  & \textbf{85.0} & 599 & --- & 0  & --- & 1 \\
                         & 10-shot & \textbf{87.9} & 563 & 38.5 & 13 & 45.8 & 24 \\
\midrule
\multirow{2}{*}{Bio.}   & 0-shot  & \textbf{72.3} & 527 & 67.2 & 64 & 33.3 & 9 \\
                         & 10-shot & \textbf{79.2} & 448 & 66.3 & 86 & 34.8 & 66 \\
\midrule
\multirow{2}{*}{Geo.}   & 0-shot  & 54.1 & 514 & \textbf{64.6} & 79 & 42.9 & 7 \\
                         & 10-shot & 71.6 & 394 & 54.4 & 114 & \textbf{73.9} & 92 \\
\bottomrule
\end{tabular}
}
\caption{Per-strategy accuracy and trial counts. The accuracy consequence of displacement depends on both the magnitude of the shift and how the alternative strategy compares to the displaced $\mathcal{S}_{\text{know}}$.}
\label{tab:per_strategy}
\end{table}

Table~\ref{tab:per_strategy} decomposes accuracy by strategy and condition. On the surface, the five domains in Table~\ref{tab:domain} tell contradictory stories: examples hurt Economics and appear to help Geoscience. The per-strategy decomposition resolves this apparent contradiction. The accuracy consequence of displacement depends not on whether displacement occurs, but on both the quality gap between the strategy that is displaced and the strategy that replaces it, and the magnitude of the shift itself. Three qualitatively distinct regimes emerge. 

\textbf{Displacement of a strong prior: accuracy drops (Economics).} Knowledge displacement is the largest here (20pp; Table~\ref{tab:strategy}), and it removes the model's best strategy. Under zero-shot, $\mathcal{S}_{\text{know}}$ achieves 88.3\% accuracy and accounts for all 600 trials. Under 10-shot, trials shift to empirical (41.7\%) and hybrid (50.9\%), lowering accuracy. The model has not gained an alternative competence; it has lost access to one it already had. Figure~\ref{fig:case_study}A illustrates the failure mode: the model replaces a correct Cournot derivation with a fitted heuristic that is numerically unstable across trials, producing the largest accuracy drop ($-$15.7pp).
 
\textbf{Displacement of a weak prior: accuracy appears to rise (Geoscience).} Displacement is similarly large (20pp), but here $\mathcal{S}_{\text{know}}$ is the \emph{weaker} strategy (54.1\%): the model recalls the correct formula but applies it with the wrong parameter convention (Figure~\ref{fig:case_study}B).  Because the 10-shot examples provide bare input-output pairs without intermediate derivations, they invite interpolation rather than formula retrieval. The headline number improves, but the improvement does not reflect better scientific reasoning. Pure empirical accuracy actually \emph{drops} (64.6\%~$\to$~54.4\%), and 32\% of correct 10-shot answers reach the right number without correct derivation. The model has traded derivation for interpolation; the accuracy gain masks, rather than refutes, displacement. A two-axis re-classification of the traces (correct-law $\times$ correct-numerics; Appendix~\ref{app:two_axis}) confirms this reading: the gain follows from a weak prior being replaced by example-supplied conventions, not from ICL improving stored knowledge.
 
\textbf{Displacement without a visible accuracy signal (Physics, Biology).} These domains show negligible accuracy change, but for structurally different reasons. In Physics, displacement itself is relatively smaller (6pp; Table~\ref{tab:strategy}). In Biology, displacement is substantial (13pp), but $\mathcal{S}_{\text{emp}}$ achieves comparable accuracy to $\mathcal{S}_{\text{know}}$ (66.3\% vs.\ 79.2\%), so the strategy substitution leaves headline numbers unchanged. Both cases illustrate what we consider the most practically dangerous regime: displacement that is invisible to any evaluation based on end-to-end accuracy alone.

\begin{figure*}[t]
\centering
\setlength{\tabcolsep}{3pt}
{\small
\begin{tabular}{p{0.47\textwidth}|p{0.47\textwidth}}
\toprule
\multicolumn{2}{l}{\textbf{Case A: ICL Displaces Correct Reasoning} --- \textsc{Cournot Consumer Surplus} (Economics)} \\
\midrule
\textbf{Zero-Shot (Parametric, Correct)} & \textbf{10-Shot (Hybrid, Wrong)} \\
\midrule
\texttt{Step 1:} \colorbox{lightgreen}{Identify Cournot model} &
\texttt{Step 1:} \colorbox{lightgreen}{Derive equilibrium quantity} \\[2pt]
``Two firms, linear demand $P = a - bQ$'' &
$Q^* = \frac{2(a-c)}{3b} = 66.7$ \\[2pt]
\texttt{Step 2:} \colorbox{lightgreen}{Derive equilibrium} &
\texttt{Step 2:} \colorbox{lightred}{Fit CS against $Q$ from examples} \\[2pt]
$q_i^* = \frac{a - c}{3b}$, $Q^* = \frac{2(a-c)}{3b}$ &
``CS grows with $Q$; nearest pairs: \\[2pt]
\texttt{Step 3:} \colorbox{lightgreen}{Compute CS from formula} &
$(Q{=}60 \to 1216.5)$, $(Q{=}70 \to 1706.5)$'' \\[4pt]
$CS = \frac{1}{2}(a - P^*)Q^* = \frac{2(a-c)^2}{9b}$ &
\texttt{Step 3:} \colorbox{lightred}{Interpolate linearly} \\[2pt]
$= \mathbf{1481.5}$ \checkmark &
$CS \approx 1216.5 + 0.67 \times 490.0 = \mathbf{1544.8}$ \ding{55} \\[2pt]
\small{\emph{Derives} from Cournot equilibrium.} &
\small{Keeps the equilibrium; \emph{interpolates} the surplus.} \\
\midrule
\multicolumn{2}{l}{\textbf{Case B: ICL Masks Displacement} --- \textsc{Sedimentology Settling} (Geoscience)} \\
\midrule
\textbf{Zero-Shot (Parametric, Wrong)} & \textbf{10-Shot (Empirical, Correct)} \\
\midrule
\texttt{Step 1:} \colorbox{lightgreen}{Derive from constant settling velocity} &
\texttt{Step 1:} \colorbox{lightred}{Infer relationship from examples} \\[2pt]
$v = H/t_f$ &
``The outputs scale with effective fall distance:'' \\[2pt]
 &
$t_{\text{new}} = t_f \cdot (H + \text{depth\_shift})/H$ \\[4pt]
\texttt{Step 2:} \colorbox{lightgreen}{Wrong shift convention} &
\texttt{Step 2:} \colorbox{lightred}{Fit from examples} \\[2pt]
$H_{\text{new}} = H - \text{depth\_shift}$ &
$(H + \text{depth\_shift})/H = 1.495919$ \\[2pt]
$t_{\text{new}} = 5.369451/0.0806076$ &
$t_{\text{new}} = 132.177438 \times 1.495919$ \\[2pt]
$\;= \mathbf{66.61}$ \ding{55} &
$\;= \mathbf{197.73}$ \checkmark \\[2pt]
\small{Right law, \textbf{wrong shift direction}.} &
\small{No physics, \textbf{correct via pattern-fitting}.} \\
\bottomrule
\end{tabular}
}
\caption{\textbf{Case A} (Economics): The model correctly derives consumer surplus from Cournot equilibrium under zero-shot, but under 10-shot fits a numerical heuristic from examples that is inconsistent across trials. \textbf{Case B} (Geoscience): The model recalls the correct law but applies the wrong shift convention; 10-shot replaces derivation with pattern-fitting that produces the right number.}
\label{fig:case_study}
\end{figure*}

\subsection{Why is knowledge recall so easy to suppress?}
\label{subsec:decoy}

The strategy competition hypothesis implies that $\mathcal{S}_{\text{know}}$ is vocabulary-triggered: the model identifies the governing equation from domain keywords, not from mathematical analysis of the inputs. If so, removing domain vocabulary should suppress knowledge recall entirely in the zero-shot condition, while 10-shot, which supplies numerical input-output patterns independent of vocabulary, should partially survive.

Table~\ref{tab:decoy} confirms the prediction. Zero-shot accuracy collapses from 55--88\% to 2.5--14\% when domain vocabulary is replaced with pseudonyms, while the mathematical structure is preserved identically. Knowledge recall depends on surface cues, not on the underlying mathematical structure of the problem: it is \emph{recognition}, not \emph{reasoning}. Under decoy conditions, 10-shot rescues performance partially (22--40\%), confirming that $\mathcal{S}_{\text{emp}}$ extracts structure from numerical patterns independently of the vocabulary channel. 

\begin{table}[t]
\centering
\begin{tabular}{lcccc}
\toprule
& \multicolumn{2}{c}{\textbf{Original}} & \multicolumn{2}{c}{\textbf{Decoy}} \\
\cmidrule(lr){2-3} \cmidrule(lr){4-5}
\textbf{Domain} & \textbf{0-shot} & \textbf{10-shot} & \textbf{0-shot} & \textbf{10-shot} \\
\midrule
Biology    & 71.2\% & 72.5\% &  8.3\% & 21.7\% \\
Chemistry  & 66.0\% & 59.8\% & 13.8\% & 25.8\% \\
Economics  & 88.3\% & 72.7\% &  2.5\% & 40.0\% \\
Geoscience & 55.3\% & 68.7\% &  3.5\% & 32.2\% \\
Physics    & 85.0\% & 85.2\% & 10.2\% & 26.2\% \\
\bottomrule
\end{tabular}
\caption{Semantic decoy: removing domain vocabulary collapses zero-shot accuracy to 2.5--14\%, while 10-shot partially rescues to 22--40\%.}
\label{tab:decoy}
\end{table}

This finding connects the three results of this section. $\mathcal{S}_{\text{know}}$ is easily displaced because it is recognition-based: triggered by domain keywords, it offers no resistance to an alternative computation pathway that examples activate. 
$\mathcal{S}_{\text{emp}}$, by contrast, operates on numerical patterns and is therefore insensitive to the vocabulary channel. When examples are present, they activate $\mathcal{S}_{\text{emp}}$ regardless of whether the model's stored formula would have served better, explaining both the direction-consistency of displacement documented in \S\ref{subsec:strategy_shift} and the domain-dependent accuracy consequences documented in \S\ref{subsec:accuracy_effects}.

\paragraph{Can structural prompting prevent displacement?} A natural objection is that practitioners would use structural prompting rather than naive ICL. Prior work has shown that ICL performance is sensitive to example selection \citep{liu2022makes}, example ordering \citep{lu2022fantastically}, and calibration \citep{zhao2021calibrate}, suggesting that careful prompt design might mitigate displacement. We tested four interventions, including explicitly naming the formula (Hint Prompting) and forcing formula declaration before seeing examples (Two-Stage), and none universally prevents displacement (Appendix~\ref{app:interventions}). No single intervention is optimal across domains, creating a practitioner's dilemma: the best strategy depends on task-specific factors, including the magnitude of knowledge displacement and the quality gap between strategies, which is precisely what the practitioner lacks access to.

\section{Implications and limitations}
\label{sec:discussion}

\paragraph{Summary.} Three results support the strategy competition account, on which ICL shifts computation from $\mathcal{S}_{\text{know}}$ toward $\mathcal{S}_{\text{emp}}$. \emph{First}, adding examples reduces the share of $\mathcal{S}_{\text{know}}$ trajectories by 6--20pp in every domain (Table~\ref{tab:strategy}), including the domains where accuracy barely moves; the model does not abandon knowledge recall selectively, it abandons it everywhere. \emph{Second}, the per-strategy decomposition (\S\ref{subsec:accuracy_effects}) explains why a single shift produces drops, apparent gains, and null results, depending on how the replacing strategy compares to the one displaced. \emph{Third}, the semantic decoy (Table~\ref{tab:decoy}) explains the vulnerability: $\mathcal{S}_{\text{know}}$ is triggered by domain vocabulary rather than mathematical structure, so it offers no resistance when examples activate an alternative pathway. The model's scientific knowledge is real but fragile --- recalled through surface cues, overridden by the mere presence of numerical examples, and displaced regardless of whether displacement helps or hurts.

\begin{figure}[t]
\centering
\includegraphics[width=0.58\linewidth]{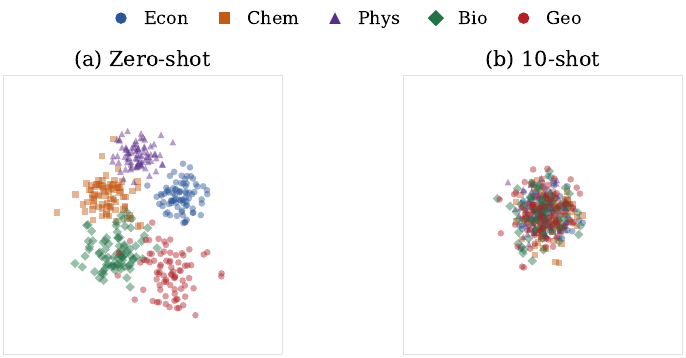}
\caption{Task-level representations in hidden-state space (Qwen2.5-7B, final layer, last token). Each point is one task, projected by classical MDS on pairwise task distances. Under 10-shot, domains overlap more, with Geoscience (red) showing the largest reduction in distance to the other domains (Table~\ref{tab:convergence_multi}).}
\label{fig:cluster}
\end{figure}

\paragraph{Internal evidence.} The behavioral evidence above is based on strategy classification of generated text. To test whether displacement is also visible in the model's internal representations, we extracted hidden states from three open-source models --- Qwen2.5-7B, Llama-3.1-8B, and Mistral-7B --- across all five domains under both conditions (Appendix~\ref{app:representational}). Under zero-shot, each domain occupies a distinct region of representation space, consistent with each domain activating its own $\mathcal{S}_{\text{know}}$ circuit. Under 10-shot, cross-domain distances shrink: in all three models, the majority of domain pairs converge, with Geoscience, the domain with the largest $\mathcal{S}_{\text{know}}$ drop in Table~\ref{tab:strategy}, converging the most toward other domains (Figure~\ref{fig:cluster}). Within-domain cosine similarity tells the same story: across all three models, Geoscience is the only domain where representations become \emph{more} uniform under 10-shot, while all other domains show decreases. The picture is consistent with $\mathcal{S}_{\text{emp}}$ being domain-agnostic: pattern-fitting from numerical examples activates similar computations regardless of the underlying domain, erasing the representational diversity that domain-specific knowledge creates.

\paragraph{Why this matters beyond knowledge recall.} The tasks in our benchmark are not arbitrary mathematical puzzles. Extracting hidden parameters from data and computing predictions through domain laws is the core computational pattern of quantitative science, from fitting kinetic models in drug development to estimating material properties from deformation curves. Prior work studies what happens when context \emph{conflicts} with the model's priors, such as flipped labels \citep{min2022rethinking,wei2023larger} and counterfactual demonstrations \citep{wu2024reasoning}. In those settings, suppression is expected. Our setting is different: the few-shot examples are fully consistent with the domain law the model already knows, yet knowledge displacement occurs nonetheless. This suggests that displacement is not triggered by conflicting signals but by the presence of numerical examples regardless of whether the patterns agree or disagree with its knowledge. In this sense, knowledge displacement may be an instance of the broader shortcut learning phenomenon \citep{geirhos2020shortcut,patel2021nlp,vafa2025world}, where models default to the computationally easiest strategy available rather than the most principled one.

\paragraph{Limitations and future work.}
Our evidence for the strategy competition hypothesis is behavioral: it rests on automated classification of chain-of-thought traces rather than direct observation of model internals. The classifier agrees with human annotators on 96\% of traces (Appendix~\ref{app:validation}), but it categorizes the model's \emph{output} text, which may not faithfully reflect the underlying computation \citep{turpin2023language, lanham2023measuring}. For this reason, the claim does not rest on trace classification alone: the semantic decoy (Table~\ref{tab:decoy}), the paired logit-difference analysis (Table~\ref{tab:logit_diff}), and the representational results (Appendix~\ref{app:representational}) are all classifier-independent. What these converging lines of evidence cannot supply is a mechanism: causal interventions such as activation patching \citep{meng2022locating,zhang2024patching} and steering-vector experiments would establish whether displacement corresponds to identifiable circuits or directions in activation space. 

Benchmark construction is deliberate: 12 canonical tasks per domain, built on textbook formulas that LLMs are likely to have encountered during pretraining. This limits generalization to messier real-world workflows. Our claim is correspondingly scoped to same-formula numerical input-output demonstrations: the worked-demonstration control (Appendix~\ref{app:worked_control}) shows that displacement persists when demonstrations carry full derivations, though the bare format may still contribute to its magnitude. Extending the benchmark to domain-specific heuristics, empirical correlations, and multi-step workflows would test whether displacement is a general property of how LLMs handle scientific reasoning under ICL. Whether substantially scaled test-time compute would eliminate displacement also remains open \citep{snell2025scaling}.

\newpage
\section*{Acknowledgments}
This work was supported by the National Research Foundation of Korea (NRF) grant (Grant No. RS-2025-00563196) and the Institute of Information \& Communications Technology Planning \& Evaluation (IITP) grants (Grant Nos. RS-2019-II191126 and RS-2024-0045974) funded by the Korean government (MSIT). This work was also supported by the AI Science Hub Program of the Ministry of Science and ICT (Grant No. N10260126).

\newpage
\appendix
 
\FloatBarrier
\section{Task inventory}
\label{app:full_inventory}

Table~\ref{tab:all_tasks} lists all 60 tasks: the question posed to the model, the
law that governs the answer, and the form of the latent-to-target computation.

\paragraph{Structural summary.} \textbf{Observable-to-latent forms} (8): fraction (29), logarithmic (8), quadratic (8), log-exponential (7), exponential (3), linear system (3), trigonometric (1), numerical root-finding (1). \textbf{Latent-to-target forms} (12): fraction (23), exponential (7), logarithmic (7), power (5), square root (4), affine (4), quadratic (3), power-fraction (2), trigonometric (2), scaled identity (1), reciprocal-log (1), log-exponential (1). \textbf{Latent variables per task}: 1 (22 tasks), 2 (24), 3 (11), 4 (3). \textbf{Observables per task}: 3--8, mode 6 (19 tasks). Fraction dominates Economics (9/12) and Geoscience (10/12).
{
\setlength{\tabcolsep}{4pt}
\begin{longtable}{>{\small\raggedright\arraybackslash}p{0.47\textwidth}>{\small\raggedright\arraybackslash}p{0.36\textwidth}>{\small}l}
\toprule
\textbf{Task} & \textbf{Governing law} & \textbf{L$\to$T} \\
\midrule
\endfirsthead
\multicolumn{3}{l}{\small\emph{Table \thetable{} continued from previous page}} \\
\toprule
\textbf{Task} & \textbf{Governing law} & \textbf{L$\to$T} \\
\midrule
\endhead
\midrule
\multicolumn{3}{r}{\small\emph{continued on next page}} \\
\endfoot
\bottomrule
\noalign{\vskip 10pt}
\caption{All 60 tasks. ``Governing law'' is the equation supplied to the two-axis
classifier; L$\to$T is the form of the latent-to-target computation.}
\label{tab:all_tasks}
\endlastfoot
\multicolumn{3}{l}{\textbf{Biology}} \\*
\midrule
Heterozygote frequency after the allele shift & \texttt{2p(1-p)} & quad \\
Metabolic rate at a new body mass & \texttt{R = a*M\^{}b} & pow \\
O2 pressure for 75\% saturation & \texttt{Y = pO2\^{}n / (K\_d + pO2\^{}n)} & pow \\
Saturation at O2 = 50 & \texttt{Y = pO2\^{}n / (K\_d + pO2\^{}n)} & p-frac \\
Substrate concentration at 80\% of Vmax & \texttt{v = Vmax*[S]/(Km+[S])} & ident \\
Elimination half-life & \texttt{t\_half = ln(2)/k} & frac \\
Slope of the binding curve at O2 = 50 & \texttt{dY/dpO2 = n*K\_d*pO2\^{}(n-1) / (K\_d + pO2\^{}n)\^{}2} & p-frac \\
Reaction rate at [S] = 200 & \texttt{v = Vmax*[S]/(Km+[S])} & frac \\
Catalytic efficiency & \texttt{Vmax/Km} & frac \\
Drug concentration at t = 24 h & \texttt{C(t) = C0*exp(-kt)} & exp \\
IC50 of the inhibitor & \texttt{IC50 = Ki*(1+[S]/Km)} & frac \\
Time to 90\% of carrying capacity & \texttt{P(t) = K/(1+A*exp(-rt))} & log \\
\midrule
\multicolumn{3}{l}{\textbf{Chemistry}} \\*
\midrule
Temperature at which the rate constant triples & \texttt{ln(k2/k1) = -(Ea/R)(1/T2-1/T1)} & r-log \\
Reaction quotient for a 0.05 V drop & \texttt{E = E0 - (RT/nF)*ln(Q)} & exp \\
New pH after adding 0.01 mol HCl & \texttt{pH = pKa + log([A-]/[HA])} & log \\
New pH after adding 0.01 mol NaOH & \texttt{pH = pKa + log([A-]/[HA])} & log \\
Half-life of the isotope & \texttt{t\_half = ln(2)/lambda} & frac \\
New [B] after [A] falls by 0.5 M & \texttt{Kc = [B]\^{}2/[A]} & sqrt \\
New [B] after [A] rises by 1 M & \texttt{Kc = [B]\^{}2/[A]} & sqrt \\
New pressure after +50 K at fixed volume & \texttt{PV = nRT} & frac \\
New volume after +50 K at fixed pressure & \texttt{PV = nRT} & frac \\
Amount remaining at t = 100 & \texttt{N(t) = N0*exp(-lambda*t)} & exp \\
Overpotential at the target current density & \texttt{eta = b*ln(j/j0)} & log \\
Peak concentration of intermediate B & \texttt{B\_max = A0*(k1/k2)\^{}(k2/(k2-k1))} & exp \\
\midrule
\multicolumn{3}{l}{\textbf{Economics}} \\*
\midrule
Deadweight loss from the tax & \texttt{DWL = 0.5*t*(Q\_pre - Q\_post)} & affine \\
Consumer surplus at the monopoly price & \texttt{CS = 0.5*(a-P)*Q} & frac \\
Deadweight loss from a \$2 per-unit tax & \texttt{DWL = 0.5*t\^{}2/(b+d)} & frac \\
New equilibrium price after demand shifts up by 5 & \texttt{P = (a*d+b*c)/(b+d)} & frac \\
Income needed for utility = 100 & \texttt{m = U*(px/alpha)\^{}alpha * (py/(1-alpha))\^{}(1-alpha)} & pow \\
Good X demand when its price doubles & \texttt{x = alpha*M/px} & frac \\
Monopoly deadweight loss & \texttt{DWL = 0.5*(P-MC)*(Qc-Qm)} & frac \\
New output after labor rises 10\% & \texttt{Y = A*L\^{}alpha*K\^{}beta} & pow \\
New marginal product after labor rises 10\% & \texttt{MPL = alpha*Y/L} & pow \\
Consumer surplus at the Cournot equilibrium & \texttt{CS = 0.5*(a-P)*Q} & frac \\
Cartel price after marginal cost rises by 3 & \texttt{P = (a+c)/2} & frac \\
New price ratio after endowment X doubles & \texttt{px/py} from Walrasian market clearing & frac \\
\midrule
\multicolumn{3}{l}{\textbf{Geoscience}} \\*
\midrule
New salinity after the conductivity shift & \texttt{S = (C-C0)/a\_c} & affine \\
Total fault offset after additional time & \texttt{v = offset/time} & affine \\
Daughter-to-parent ratio after added time & \texttt{R = exp(lambda*t) - 1} & exp \\
Temperature at the target depth & \texttt{T(z) = T0 + G*z} & affine \\
New basal shear stress after the thickness shift & \texttt{tau = rho*g*H*sin(alpha)} & trig \\
New discharge after the head shift & \texttt{Q = K*A*(dh/L)} & frac \\
New fall time after the depth shift & \texttt{w\_s = (rho\_p - rho\_f)*g*d\^{}2/(18*mu)} & frac \\
New S-P lag after the distance shift & \texttt{dt = D*(1/vs - 1/vp)} & frac \\
New surface uplift after the pressure shift & \texttt{u = beta*dP/depth} & frac \\
New dew point after the temperature shift & \texttt{es = 6.112*exp(17.67*T/(T+243.5))}, inverted & lg-ex \\
Drawdown at a new radial distance & \texttt{s = Q/(4*pi*T)*ln(2.25*T*t/(r\^{}2*S))} & log \\
Melt composition at a lower temperature & \texttt{CL = C0*F\^{}(D-1)} & exp \\
\midrule
\multicolumn{3}{l}{\textbf{Physics}} \\*
\midrule
Refracted angle at a new incident angle & \texttt{n1*sin(theta1) = n2*sin(theta2)} & trig \\
Image distance at a new object distance & \texttt{1/f = 1/do + 1/di} & frac \\
Position at t = 20 & \texttt{x(t) = x0 + v0*t + 0.5*a*t\^{}2} & quad \\
Time to reach velocity 50 & \texttt{v = v0 + a*t} & frac \\
Range on the Moon ($g = 1.62$\,m/s\textsuperscript{2}) & \texttt{R = 2*vx*vy/g} & frac \\
Initial kinetic energy per kg & \texttt{KE/m = 0.5*(vx\^{}2+vy\^{}2)} & quad \\
Capacitor voltage at time 2*t2 & \texttt{V(t) = Vmax*(1-exp(-t/tau))} & exp \\
Time to 90\% of full charge & \texttt{t = -tau*ln(0.1)} & log \\
Maximum speed of the oscillator & \texttt{v\_max = A*sqrt(k/m)} & sqrt \\
New period when the mass doubles & \texttt{T = 2*pi*sqrt(m/k)} & sqrt \\
New acceleration after m1 rises 20\% & \texttt{a = (m1-m2)*g/(m1+m2+I/R\^{}2)} & frac \\
Time to cool to 30$^\circ$C & \texttt{T(t) = T\_env + A*exp(-k*t)} & log \\
\end{longtable}
}

\FloatBarrier
\section{Benchmark difficulty balance}
\label{app:tasks}
 
\textbf{Domain balance}: Kruskal-Wallis $p = 0.44$. Difficulty tiers (Trivial/Easy/Moderate/Hard/Very Hard) are distributed similarly across all five domains, with most tasks in the Moderate tier.
 
\paragraph{Difficulty standardization.}
To ensure the inversion is not an artifact of unequal task difficulty across domains, we standardized difficulty using four structural criteria: (1) number of observables, (2) number of latent variables, (3) observable-to-latent equation format complexity, and (4) latent-to-target equation format complexity. The sum is mapped to ordinal tiers.
 
\begin{table}[H]
\centering
\small
\begin{tabular}{p{0.34\linewidth}p{0.58\linewidth}}
\toprule
\textbf{Component} & \textbf{Definition / Role} \\
\midrule
Number of observables & How many surface variables are provided. More observables increase extraction difficulty. \\
Number of latent variables & How many hidden parameters must be recovered before the target can be computed. \\
Observable $\rightarrow$ latent format & Structural form required to infer latents from observables (e.g., linear, fraction, logarithmic, quadratic, trigonometric). \\
Latent $\rightarrow$ target format & Structural form of the final computation once latents are recovered. \\
\midrule
Total score & Sum of the four components (0--11). \\
Tier mapping & 0--2: Trivial;\; 3--4: Easy;\; 5--6: Moderate;\; 7--8: Hard;\; 9--11: Very Hard. \\
\bottomrule
\end{tabular}
\caption{Difficulty benchmark based on four structural criteria.}
\label{tab:benchmark_4criteria}
\end{table}
 
\begin{table}[H]
\centering
\small
\begin{tabular}{lcccccc}
\toprule
\textbf{Domain} & \textbf{\# Tasks} & \textbf{Trivial} & \textbf{Easy} & \textbf{Moderate} & \textbf{Hard} & \textbf{Very Hard} \\
\midrule
Biology    & 12 & 0 & 2 & 8 & 1 & 1 \\
Chemistry  & 12 & 0 & 4 & 6 & 1 & 1 \\
Economics  & 12 & 1 & 0 & 8 & 2 & 1 \\
Geoscience & 12 & 0 & 2 & 8 & 1 & 1 \\
Physics    & 12 & 1 & 1 & 8 & 1 & 1 \\
\bottomrule
\end{tabular}
\caption{Tier distribution by domain. Most tasks in every domain fall into the Moderate tier (Kruskal-Wallis $p = 0.44$).}
\label{tab:domain_tier_split}
\end{table}

\FloatBarrier
\section{Classifier validation}
\label{app:validation}
 
\begin{table}[H]
\centering
{\small
\begin{tabular}{llcccc}
\toprule
\textbf{Domain} & \textbf{Cond.} & \textbf{Param.} & \textbf{Empir.} & \textbf{Hybrid} & \textbf{Total} \\
\midrule
\multirow{2}{*}{Economics}  & 0-shot  & 600 & 0 & 0 & 600 \\
                            & 10-shot & 482 & 12 & 106 & 600 \\
\midrule
\multirow{2}{*}{Chemistry}  & 0-shot  & 551 & 46 & 3 & 600 \\
                            & 10-shot & 488 & 78 & 34 & 600 \\
\midrule
\multirow{2}{*}{Physics}    & 0-shot  & 599 & 0 & 1 & 600 \\
                            & 10-shot & 563 & 13 & 24 & 600 \\
\midrule
\multirow{2}{*}{Biology}    & 0-shot  & 527 & 64 & 9 & 600 \\
                            & 10-shot & 448 & 86 & 66 & 600 \\
\midrule
\multirow{2}{*}{Geoscience} & 0-shot  & 514 & 79 & 7 & 600 \\
                            & 10-shot & 394 & 114 & 92 & 600 \\
\bottomrule
\end{tabular}
}
\caption{Raw strategy classification counts.}
\label{tab:strategy_counts}
\end{table}
 
\textsc{Parametric} trajectories reference named equations and derive parameters algebraically; \textsc{Empirical} trajectories fit models to numerical extrapolation from examples; \textsc{Hybrid} trajectories invoke a formula but calibrate coefficients from examples. The two case studies in Figure~\ref{fig:case_study} illustrate each category.
 
\paragraph{Human annotation validation.} Two independent annotators each labeled a stratified subset of 100 CoT traces (20 per domain, evenly split between zero-shot and 10-shot conditions) using the same three-category scheme. Inter-annotator agreement was 98.0\% (98/100 traces). The 2 disagreements involved the Parametric/Hybrid boundary (whether coefficient calibration from examples was present); neither involved the Parametric/Empirical distinction that drives our main results. Agreement between the human consensus labels and GPT-4o's automated classifications was 96.0\% (96/100), confirming that the strategy categories are well-defined and that the automated classifier does not introduce systematic bias in the direction of the observed shift.

\FloatBarrier
\section{Two-axis strategy decomposition: correct-law $\times$ correct-numerics}
\label{app:two_axis}

The three-class taxonomy collapses two properties of a trace: whether the model invoked the correct governing law, and whether it used correct numerical values (recovered parameters, constants, arithmetic). We re-classified traces along these two binary axes with the same automated protocol (Appendix~\ref{app:implementation}). Table~\ref{tab:two_axis_geo} reports Geoscience, whose accuracy gain the decomposition disambiguates; Table~\ref{tab:two_axis_delta} reports the 0-shot to 10-shot shifts for all five domains.

\begin{table}[H]
\centering
{\small
\begin{tabular}{lcccc}
\toprule
\textbf{Geoscience (\% trials)} & \textbf{CL+CN} & \textbf{CL+WN} & \textbf{WL+CN} & \textbf{WL+WN} \\
\midrule
0-shot  & 43 & 44 & 10 & 3 \\
10-shot & 47 & 20 & 21 & 12 \\
$\Delta$ & $+$4 & $-$24 & $+$11 & $+$9 \\
\bottomrule
\end{tabular}
}
\caption{Two-axis decomposition for Geoscience. The accuracy gain comes from two shifts: correct-law-wrong-numerics trials drop sharply, and wrong-law-correct-numerics trials rise (more example-driven trials landing on the right number, as in Figure~\ref{fig:case_study}B).}
\label{tab:two_axis_geo}
\end{table}

\begin{table}[H]
\centering
{\small
\begin{tabular}{lccccc}
\toprule
\textbf{Domain} & $\boldsymbol{\Delta}$\textbf{CL+CN} & $\boldsymbol{\Delta}$\textbf{CL+WN} & $\boldsymbol{\Delta}$\textbf{WL+CN} & $\boldsymbol{\Delta}$\textbf{WL+WN} & $\boldsymbol{\Delta}$\textbf{CL total} \\
\midrule
Economics  & $-$23 & $+$7  & $+$7  & $+$7 & $-$16 \\
Chemistry  & $-$12 & $+$1  & $+$6  & $+$4 & $-$11 \\
Physics    & $-$5  & $-$1  & $+$3  & $+$1 & $-$6 \\
Biology    & $-$5  & $-$9  & $+$7  & $+$6 & $-$14 \\
Geoscience & $+$4  & $-$24 & $+$11 & $+$9 & $-$20 \\
\bottomrule
\end{tabular}
}
\caption{Two-axis shifts (10-shot minus 0-shot, pp of trials) for all five domains.}
\label{tab:two_axis_delta}
\end{table}

The correct-law share drops in every domain (6 to 20pp) while the wrong-law share grows, matching the direction of the Parametric-share drops in Table~\ref{tab:strategy}. Geoscience is where the largest shift to wrong-law-correct-numerics cell occurs, because its zero-shot parametric accuracy was unusually low (54.1\%, Table~\ref{tab:per_strategy}): the $+$13.3pp gain follows from a weak prior being replaced by example-supplied conventions, not from ICL improving stored knowledge. Elsewhere the same shift adds wrong trials (Economics: $\Delta$CL+CN of $-$23pp).

\FloatBarrier
\section{Full per-task results}
\label{app:per_task}
 
{
\setlength{\tabcolsep}{2.5pt}
\begin{longtable}{>{\small}l>{\small}c>{\small}c>{\small}c>{\small}c}
\toprule
\textbf{Task} & \textbf{O$\to$L} & \textbf{0-shot} & \textbf{10-shot} & $\boldsymbol{\Delta}$ \\
\midrule
\endfirsthead
\multicolumn{5}{l}{\small\emph{Table~\ref{tab:per_task_full} continued from previous page}} \\
\toprule
\textbf{Task} & \textbf{O$\to$L} & \textbf{0-shot} & \textbf{10-shot} & $\boldsymbol{\Delta}$ \\
\midrule
\endhead
\endfoot
\midrule
\multicolumn{5}{r}{\small\emph{continued on next page}} \\
\endfoot
\bottomrule
\noalign{\vskip 10pt}
\caption{Per-task accuracy (50 trials). Colored $\Delta$: $|\Delta| > 10$pp.}
\label{tab:per_task_full}
\endlastfoot
B: Allometry Metabolic rate & log & 68\% & 70\% & +2 \\
B: CompetitiveInhibition IC50 & frac & 48\% & 38\% & $-$10 \\
B: HardyWeinberg Heterozygote & frac & 2\% & 18\% & \cellcolor{lightgreen}+16 \\
B: Hill O2 for 75\% sat. & log-exp & 84\% & 90\% & +6 \\
B: Hill Saturation at O2=50 & log-exp & 100\% & 100\% & +0 \\
B: Hill Slope at O2=50 & log-exp & 90\% & 94\% & +4 \\
B: LogisticGrowth 90\% of K & exp & 74\% & 64\% & $-$10 \\
B: MichaelisMenten Efficiency & frac & 80\% & 70\% & $-$10 \\
B: MichaelisMenten Rate@200 & frac & 76\% & 80\% & +4 \\
B: MichaelisMenten 80\% Vmax & frac & 32\% & 50\% & \cellcolor{lightgreen}+18 \\
B: Pharma Conc@24h & log-exp & 100\% & 96\% & $-$4 \\
B: Pharma Half life & log-exp & 100\% & 100\% & +0 \\
\midrule
C: Arrhenius 3x rate & log & 100\% & 100\% & +0 \\
C: Buffer pH+HCl & log & 100\% & 100\% & +0 \\
C: Buffer pH+NaOH & log & 100\% & 100\% & +0 \\
C: ConsecutiveRxn Peak & exp & 0\% & 16\% & \cellcolor{lightgreen}+16 \\
C: Decay Amount@100 & log-exp & 92\% & 78\% & \cellcolor{lightred}$-$14 \\
C: Decay Half life & log-exp & 100\% & 98\% & $-$2 \\
C: Equilibrium (A dec.) & quad & 2\% & 2\% & +0 \\
C: Equilibrium (A inc.) & quad & 78\% & 22\% & \cellcolor{lightred}$-$56 \\
C: IdealGas Pressure & frac & 64\% & 28\% & \cellcolor{lightred}$-$36 \\
C: IdealGas Volume & frac & 44\% & 52\% & +8 \\
C: Nernst Q & log & 100\% & 44\% & \cellcolor{lightred}$-$56 \\
C: Tafel Overpotential & frac & 12\% & 78\% & \cellcolor{lightgreen}+66 \\
\midrule
E: CobbDouglas MPL & frac & 100\% & 96\% & $-$4 \\
E: CobbDouglas Output & frac & 100\% & 100\% & +0 \\
E: Competitive DWL & quad & 98\% & 64\% & \cellcolor{lightred}$-$34 \\
E: Competitive Price & quad & 96\% & 92\% & $-$4 \\
E: Consumer Income & frac & 76\% & 14\% & \cellcolor{lightred}$-$62 \\
E: Consumer GoodX & frac & 100\% & 100\% & +0 \\
E: Cournot CS & frac & 100\% & 74\% & \cellcolor{lightred}$-$26 \\
E: Cournot Cartel & frac & 90\% & 58\% & \cellcolor{lightred}$-$32 \\
E: Edgeworth Eq. & frac & 0\% & 0\% & +0 \\
E: Monopoly CS & frac & 100\% & 98\% & $-$2 \\
E: Monopoly DWL & frac & 100\% & 80\% & \cellcolor{lightred}$-$20 \\
E: Tax DWL & lin & 100\% & 96\% & $-$4 \\
\midrule
G: Aquifer Drawdown & frac & 32\% & 46\% & +14 \\
G: Geochronology & log & 56\% & 32\% & \cellcolor{lightred}$-$24 \\
G: Geodesy Slip rate & frac & 58\% & 98\% & \cellcolor{lightgreen}+40 \\
G: Geothermics Temp & frac & 100\% & 96\% & $-$4 \\
G: Glaciology Thickness & frac & 98\% & 92\% & $-$6 \\
G: Hydrology K & frac & 6\% & 44\% & \cellcolor{lightgreen}+38 \\
G: MagmaticDiff Comp & exp & 74\% & 76\% & +2 \\
G: Meteorology RH & frac & 100\% & 94\% & $-$6 \\
G: Oceanography Salinity & frac & 4\% & 36\% & \cellcolor{lightgreen}+32 \\
G: Sedimentology Settling & frac & 26\% & 100\% & \cellcolor{lightgreen}+74 \\
G: Seismology Distance & frac & 100\% & 100\% & +0 \\
G: Volcanology Pressure & frac & 10\% & 10\% & +0 \\
\midrule
P: Atwood Acceleration & frac & 0\% & 16\% & \cellcolor{lightgreen}+16 \\
P: Cooling Time@30$^\circ$ & root & 26\% & 18\% & $-$8 \\
P: Kinematics Pos@20 & lin & 98\% & 98\% & +0 \\
P: Kinematics Time@v50 & lin & 100\% & 100\% & +0 \\
P: Projectile Range & quad & 100\% & 100\% & +0 \\
P: Projectile KE & quad & 100\% & 100\% & +0 \\
P: RC 90\% charge & log & 100\% & 100\% & +0 \\
P: RC Voltage@2t & log & 100\% & 100\% & +0 \\
P: SHM Max speed & quad & 96\% & 98\% & +2 \\
P: SHM Period@2m & quad & 100\% & 100\% & +0 \\
P: Snell Refraction & trig & 100\% & 98\% & $-$2 \\
P: ThinLens Image & frac & 100\% & 94\% & $-$6 \\
\end{longtable}
}

 \FloatBarrier
\section{Statistical treatment}
\label{app:stats}
 
Each of the 50 trials per task uses an independently generated problem instance with different numerical values. To account for heterogeneity across tasks within each domain, we supplement the trial-level McNemar tests (Table~\ref{tab:domain}) with a DerSimonian-Laird random-effects meta-analysis \citep{dersimonian1986meta} treating each task as a study. This models both within-task sampling variance and between-task heterogeneity.
 
Under this analysis, Economics ($\hat{\mu} = -12.8$pp, $z = -4.06$, $p < .0001$; 95\% CI: $[-19.0, -6.6]$) and Geoscience ($\hat{\mu} = +13.1$pp, $z = 2.13$, $p = .033$; 95\% CI: $[+1.0, +25.2]$) remain significant despite substantial task-level heterogeneity ($I^2 = 91\%$ and $95\%$, respectively). Chemistry trends negative ($-5.3$pp, $p = .24$) but does not reach significance due to opposing task-level effects ($I^2 = 95\%$). Physics and Biology show no effect, as expected.

\FloatBarrier
\section{Cross-model numerical results}
\label{app:cross_model}

Table~\ref{tab:cross_model_full} reports the full zero-shot and 10-shot accuracies for all four models. These are the numerical values underlying Figure~\ref{fig:main_results}.

\begin{table}[H]
\centering
{\small
\begin{tabular}{llcccc}
\toprule
& & \textbf{GPT-5.2} & \textbf{Claude 4.6} & \textbf{Llama 4} & \textbf{DeepSeek V3} \\
\midrule
\multirow{3}{*}{Economics}
  & 0-shot  & 88.3 & 100.0 & 81.4 & 78.1 \\
  & 10-shot & 72.7 & 93.1  & 71.1 & 71.7 \\
  & $\Delta$ & \textbf{$-$15.7} & \textbf{$-$6.9} & \textbf{$-$10.3} & \textbf{$-$6.4} \\
\midrule
\multirow{3}{*}{Chemistry}
  & 0-shot  & 66.0 & 94.7 & 87.8 & 78.1 \\
  & 10-shot & 59.8 & 90.8 & 93.1 & 76.9 \\
  & $\Delta$ & \textbf{$-$6.2} & \textbf{$-$3.9} & \textbf{$+$5.3} & $-$1.2 \\
\midrule
\multirow{3}{*}{Physics}
  & 0-shot  & 85.0 & 94.7 & 91.9 & 91.1 \\
  & 10-shot & 85.2 & 95.6 & 89.4 & 89.7 \\
  & $\Delta$ & $+$0.2 & $+$0.9 & $-$2.5 & $-$1.4 \\
\midrule
\multirow{3}{*}{Biology}
  & 0-shot  & 71.2 & 98.3 & 72.8 & 78.9 \\
  & 10-shot & 72.5 & 98.9 & 74.7 & 80.6 \\
  & $\Delta$ & $+$1.3 & $+$0.6 & $+$1.9 & $+$1.7 \\
\midrule
\multirow{3}{*}{Geoscience}
  & 0-shot  & 55.3 & 75.0 & 63.1 & 62.5 \\
  & 10-shot & 68.7 & 90.3 & 81.4 & 72.5 \\
  & $\Delta$ & \textbf{$+$13.3} & \textbf{$+$15.3} & \textbf{$+$18.3} & \textbf{$+$10.0} \\
\bottomrule
\end{tabular}
}
\caption{Full cross-model results. Accuracy (\%) under zero-shot and 10-shot conditions (GPT-5.2: 50 trials per task; others: 30 trials per task). Bold $\Delta$ indicates $|\Delta| > 3$pp. Economics is negative in 4/4 models; Geoscience is positive in 4/4.}
\label{tab:cross_model_full}
\end{table}

\FloatBarrier
\section{Within-format confound control}
\label{app:confound}
 
\begin{table}[H]
\centering
\begin{tabular}{lcccc}
\toprule
\textbf{Domain} & $\boldsymbol{n}$ & \textbf{Zero-Shot} & \textbf{10-Shot} & $\boldsymbol{\Delta}$\textbf{(pp)} \\
\midrule
Economics (fraction)  & 9  & 85.1\% & 68.9\% & \cellcolor{lightred}$-$16.2 \\
Geoscience (fraction) & 10 & 53.4\% & 71.6\% & \cellcolor{lightgreen}+18.2 \\
\bottomrule
\end{tabular}
\caption{Same equation format, different ICL effects in Economics and Geoscience.}
\label{tab:confound}
\end{table}

\FloatBarrier
\section{Ceiling task analysis}
\label{app:ceiling_analysis}
 
Of 60 tasks, 14 achieve 100\% under both conditions (concentrating in Physics: 6/12). Restricting to non-ceiling tasks amplifies the inversion.
 
\begin{table}[H]
\centering
{\small
\begin{tabular}{lccccc}
\toprule
\textbf{Domain} & \textbf{Ceil.} & \textbf{Non-ceil.} & \textbf{0-shot} & \textbf{10-shot} & $\boldsymbol{\Delta}$\textbf{(pp)} \\
\midrule
Physics    & 6 & 6  & 70.0\% & 70.3\% & +0.3 \\
Economics  & 2 & 10 & 86.0\% & 67.2\% & \cellcolor{lightred}$-$18.8 \\
Geoscience & 1 & 11 & 51.3\% & 65.8\% & \cellcolor{lightgreen}+14.5 \\
Chemistry  & 3 & 9  & 54.7\% & 46.4\% & \cellcolor{lightred}$-$8.2 \\
Biology    & 2 & 10 & 65.4\% & 67.0\% & $+$1.6 \\
\bottomrule
\end{tabular}
}
\caption{Non-ceiling tasks: the inversion amplifies ($-$18.8pp vs.\ $+$14.5pp).}
\label{tab:ceiling_ext}
\end{table}

\FloatBarrier
\section{Sensitivity analysis}
\label{app:sensitivity}
 
\begin{table}[H]
\centering
{\small
\begin{tabular}{lccc}
\toprule
\textbf{Domain} & \textbf{Full $\Delta$} & \textbf{Trimmed $\Delta$} & \textbf{Dropped tasks} \\
\midrule
Economics  & $-$15.7pp & $-$9.2pp  & Consumer Income, Competitive DWL \\
Chemistry  & $-$6.2pp  & $-$8.4pp  & Tafel, Equilibrium \\
Geoscience & $+$13.3pp & $+$4.6pp  & Sedimentology, Geodesy \\
Biology    & $+$1.3pp  & $-$1.8pp  & MichaelisMenten, HardyWeinberg \\
Physics    & $+$0.2pp  & $-$0.6pp  & Atwood, Cooling \\
\bottomrule
\end{tabular}
}
\caption{Inversion sign preserved after dropping 2 most extreme tasks per domain; the two null domains change sign within noise.}
\label{tab:sensitivity}
\end{table}

\FloatBarrier
\section{Accuracy by equation complexity}
\label{app:linear_trap}
 
\begin{table}[H]
\centering
\begin{tabular}{lccc}
\toprule
\textbf{Equation Type} & \textbf{Zero-Shot} & \textbf{10-Shot} & $\boldsymbol{\Delta}$\textbf{(pp)} \\
\midrule
Linear O$\to$L ($n=3$) & 99.3\% & 98.0\% & $-$1.3 \\
Non-linear O$\to$L ($n=57$) & 71.8\% & 70.4\% & $-$1.4 \\
\midrule
Linear L$\to$T ($n=5$) & 58.8\% & 75.2\% & +16.4 \\
Non-linear L$\to$T ($n=55$) & 74.5\% & 71.5\% & $-$3.0 \\
\bottomrule
\end{tabular}
\caption{ICL helps linear targets ($+$16.4pp) but hurts non-linear ones ($-$3.0pp).}
\label{tab:complexity_ext}
\end{table}
 
\begin{table}[H]
\centering
\begin{tabular}{lccc}
\toprule
\textbf{O$\to$L Format} & $\boldsymbol{n}$ & \textbf{Zero-Shot} & \textbf{10-Shot} \\
\midrule
Trigonometric       & 1  & 100.0\% & 98.0\% \\
Linear system       & 3  & 99.3\% & 98.0\% \\
Log-exponential     & 7  & 95.1\% & 93.7\% \\
Logarithmic         & 8  & 90.5\% & 80.8\% \\
Quadratic           & 8  & 83.8\% & 72.2\% \\
Fraction            & 29 & 60.6\% & 64.1\% \\
Exponential         & 3  & 49.3\% & 52.0\% \\
Numerical root      & 1  & 26.0\% & 18.0\% \\
\bottomrule
\end{tabular}
\caption{Accuracy by observable-to-latent equation format (all domains pooled). Harder formats (quadratic, fraction) show more variable ICL effects.}
\label{tab:difficulty_ext}
\end{table}

\FloatBarrier
\section{Prompting interventions}
\label{app:interventions}
 
We tested whether structural prompting can prevent knowledge displacement. Four interventions were evaluated (20 trials per task):
 
\begin{table}[H]
\centering
\setlength{\tabcolsep}{3pt}
{\small
\begin{tabular}{lcccccc}
\toprule
\textbf{Domain} & \textbf{0-shot} & \textbf{10-shot} &  \textbf{Hint} & \textbf{Neg.} & \textbf{Self-Verify.} & \textbf{2-Stage} \\
\midrule
Economics  & \textbf{88.3} & 72.7 & 67.9 & 69.2 & 62.9 & 75.4 \\
Chemistry  & \textbf{66.0} & 59.8 & 51.7 & 49.6 & 52.9 & 63.3 \\
Physics    & \textbf{85.0} & 85.2 & 77.5 & 80.8 & 72.5 & 85.0 \\
Biology    & 71.2 & \textbf{72.5} & 48.3 & 48.8 & 46.3 & 68.8 \\
Geoscience & 55.3 & 68.7 & \textbf{77.9} & 76.3 & 70.4 & 70.0 \\
\bottomrule
\end{tabular}
}
\caption{Accuracy (\%) across prompting interventions (20 trials per task).}
\label{tab:interventions}
\end{table}
 
\textbf{Two-Stage prompting} explicitly declares the target formula before seeing examples. This partially recovers displacement ($+2.7$pp over 10-shot in Economics; $+3.5$pp in Chemistry) and exceeds zero-shot in Geoscience. However, in Economics, Two-Stage still falls well short of zero-shot (75.4\% vs.\ 88.3\%), confirming that even structured prompting cannot fully prevent displacement when the model's prior is already near-optimal.
 
\textbf{Hint Prompting} provides the equation name alongside the 10-shot examples. It is most effective for Geoscience ($+9.2$pp over 10-shot), suggesting that explicit formula cues can partially counteract displacement in low-prior domains. However, Hint Prompting still underperforms zero-shot in Economics ($67.9$ vs.\ $88.3$), indicating that even when the correct formula is named, the presence of examples still shifts computation away from $\mathcal{S}_{\text{know}}$.
 
\textbf{Negative Prompting} instructs the model not to fit patterns from examples. \textbf{Self-Verification} asks the model to verify its 10-shot answer by re-deriving from first principles. Neither consistently recovers zero-shot performance.
 
No single intervention is universally optimal, and the best strategy remains domain-dependent (see \S\ref{sec:discussion}).
 
\FloatBarrier
\section{Worked-demonstration control}
\label{app:worked_control}

The 10-shot prompt of the main experiments provides bare input-output pairs, raising the question of whether displacement reflects the examples or the absence of reasoning steps in this format. A same-task worked demonstration states the formula and full derivation for the test problem itself, so the model can read off the procedure directly; accuracy saturates near ceiling (98.6--99.7\% across domains; Table~\ref{tab:worked_control}), which measures procedural copying rather than knowledge retrieval. We therefore constructed cross-task worked demonstrations: each of the 10 demonstrations carries a full derivation (formula, parameter extraction, arithmetic) for a different sibling task within the same domain, so reasoning is present in every demonstration but the procedure cannot be transcribed onto the test problem.

\begin{table}[H]

\centering

\setlength{\tabcolsep}{4pt}

{\small

\begin{tabular}{lccccc}

\toprule

\textbf{Condition (GPT-5.2, 30 trials/task)} & \textbf{Econ.} & \textbf{Chem.} & \textbf{Phys.} & \textbf{Bio.} & \textbf{Geo.} \\

\midrule

Zero-shot & 88.3 & 66.0 & 85.0 & 71.2 & 55.3 \\

10-shot bare I/O & 72.7 & 59.8 & 85.2 & 72.5 & 68.7 \\

10-shot cross-task worked-CoT & 77.5 & 62.2 & 85.6 & 69.2 & 66.9 \\

10-shot same-task worked-CoT (reference) & 99.7 & 98.9 & 99.4 & 99.2 & 98.6 \\

\bottomrule

\end{tabular}

}

\caption{Worked-demonstration control. Cross-task worked demonstrations carry full reasoning but a non-transferable procedure; accuracy recovers only partially.}

\label{tab:worked_control}

\end{table}
Even with explicit reasoning in every demonstration, displacement persists: accuracy recovers only partially, short of zero-shot, matching the partial recoveries of the Appendix~\ref{app:interventions} interventions. This control shows that reasoning content does not prevent displacement, but it is not a full format-vs-examples decomposition, since cross-task demonstrations vary task and formula identity rather than format alone; a clean dissociation belongs at the mechanistic level (\S\ref{sec:discussion}).

\FloatBarrier
\section{Boundary test}
\label{app:boundary}
 
We tested whether LLMs can induce non-linear latent relationships purely from examples within the practical range of $k \leq 20$ shots. Six target functions of increasing complexity were tested across three blocks. L0 (directly observable) reaches 100\% with $\geq$5 examples. L1--L5 (requiring non-linear latent composition) plateau near 15\% at $k=20$. 10--20 shots represents the practical ceiling for most deployed systems due to context window constraints and cost \citep{agarwal2024manyshot}.
 
\begin{table}[H]
\centering
\setlength{\tabcolsep}{4pt}
{\small
\begin{tabular}{ll cccc cccc cccc}
\toprule
& & \multicolumn{4}{c}{\textbf{Block A}} & \multicolumn{4}{c}{\textbf{Block B}} & \multicolumn{4}{c}{\textbf{Block C}} \\
\cmidrule(lr){3-6} \cmidrule(lr){7-10} \cmidrule(lr){11-14}
& & 0 & 5 & 10 & 20 & 0 & 5 & 10 & 20 & 0 & 5 & 10 & 20 \\
\midrule
L0 & Obs.  & 0 & 100 & 100 & 100 & 0 & 100 & 100 & 100 & 2 & 100 & 100 & 100 \\
L1 & Sum   & 0 & 6 & 14 & 16 & 0 & 12 & 14 & 16 & 0 & 4 & 2 & 4 \\
L2 & Ratio & 0 & 8 & 8 & 8 & 0 & 10 & 8 & 2 & 2 & 6 & 28 & 20 \\
L3 & Harm. & 0 & 6 & 16 & 30 & 0 & 20 & 24 & 20 & 0 & 2 & 4 & 8 \\
L4 & Quad. & 0 & 28 & 34 & 26 & 0 & 26 & 26 & 18 & 0 & 6 & 12 & 14 \\
L5 & Exp.  & 0 & 12 & 4 & 8 & 0 & 10 & 8 & 20 & 0 & 4 & 4 & 4 \\
\bottomrule
\end{tabular}
}
\caption{Boundary test accuracy (\%, 50 trials per cell).}
\label{tab:boundary}
\end{table}

 \FloatBarrier
\section{Representational convergence analysis}
\label{app:representational}
 
To test whether knowledge displacement is visible in the model's internal representations, we extracted hidden states from three open-source models: Qwen2.5-7B-Instruct (28 layers, dim=3584), Mistral-7B-Instruct-v0.3 (32 layers, dim=4096), and Llama-3.1-8B-Instruct (32 layers, dim=4096). For each model, we ran all 12 tasks per domain with 30 instances per task under both zero-shot and 10-shot conditions (360 vectors per domain, 3,600 forward passes per model), recording the hidden state at the last token position from the final layer.
 
For each model we computed two metrics: (1) cross-domain centroid distances (Euclidean distance between mean hidden-state vectors of each domain pair, under each condition), and (2) within-domain cosine similarity (mean pairwise cosine similarity among all vectors within each domain). A distance ratio (ZS/10S) above 1.0 indicates convergence under 10-shot; an increase in cosine similarity indicates more uniform representations.

For visualization (Figure~\ref{fig:cluster}) we additionally computed the 60$\times$60 matrix of pairwise Euclidean distances between task centroids and projected it by classical MDS, fit separately per condition. Both panels are plotted on identical axes; orientation is arbitrary, so relative distances rather than absolute positions are interpretable.
 
\paragraph{Cross-domain convergence.} Table~\ref{tab:convergence_multi} reports the number of converging pairs (ratio $>$ 1.0 out of 10 total), the mean ratio, and the mean ratio for Geoscience pairs specifically.
 
\begin{table}[H]
\centering
{\small
\begin{tabular}{lccc}
\toprule
\textbf{Model} & \textbf{Pairs converge} & \textbf{Mean ratio} & \textbf{Geo mean ratio} \\
\midrule
Qwen2.5-7B & 8/10 & 1.092 & 1.19 \\
Mistral-7B & 8/10 & 1.068 & 1.14 \\
Llama-3.1-8B & 9/10 & 1.095 & 1.17 \\
\bottomrule
\end{tabular}
}
\caption{Cross-domain convergence summary. The majority of domain pairs converge under 10-shot in all three models, with Geoscience pairs showing the strongest effect.}
\label{tab:convergence_multi}
\end{table}
 
\paragraph{Within-domain cosine similarity.} Table~\ref{tab:cosine_multi} reports the change in within-domain cosine similarity (10-shot minus zero-shot) for each domain and model.
 
\begin{table}[H]
\centering
{\small
\begin{tabular}{lccc}
\toprule
\textbf{Domain} & \textbf{Qwen-7B} & \textbf{Mistral-7B} & \textbf{Llama-8B} \\
\midrule
Economics  & $-$0.008 & $-$0.006 & $-$0.009 \\
Chemistry  & $-$0.006 & $-$0.009 & $-$0.007 \\
Physics    & $-$0.007 & $-$0.007 & $-$0.008 \\
Biology    & $-$0.004 & $-$0.004 & $-$0.005 \\
Geoscience & \textbf{+0.008} & \textbf{+0.006} & \textbf{+0.009} \\
\bottomrule
\end{tabular}
}
\caption{Within-domain cosine similarity change (10S $-$ ZS). Across all three models, Geoscience is the only domain where similarity \emph{increases} under 10-shot.}
\label{tab:cosine_multi}
\end{table}
 
\begin{table}[H]
\centering
{\small
\begin{tabular}{lccc}
\toprule
\textbf{Domain pair} & \textbf{ZS dist.} & \textbf{10S dist.} & \textbf{Ratio} \\
\midrule
Econ $\leftrightarrow$ Chem & 54.1 & 56.3 & 0.96 \\
Econ $\leftrightarrow$ Phys & 59.7 & 55.2 & 1.08 \\
Econ $\leftrightarrow$ Bio & 60.3 & 61.8 & 0.98 \\
Econ $\leftrightarrow$ Geo & 73.8 & 63.4 & 1.16 \\
Chem $\leftrightarrow$ Phys & 52.1 & 51.4 & 1.01 \\
Chem $\leftrightarrow$ Bio & 55.9 & 52.6 & 1.06 \\
Chem $\leftrightarrow$ Geo & 73.2 & 58.1 & \textbf{1.26} \\
Phys $\leftrightarrow$ Bio & 70.4 & 66.8 & 1.05 \\
Phys $\leftrightarrow$ Geo & 85.1 & 69.4 & \textbf{1.23} \\
Bio $\leftrightarrow$ Geo & 56.2 & 50.3 & 1.12 \\
\bottomrule
\end{tabular}
}
\caption{Cross-domain centroid distances, Qwen2.5-7B (12 tasks $\times$ 30 instances).}
\label{tab:convergence_qwen}
\end{table}
 
\begin{table}[H]
\centering
{\small
\begin{tabular}{lccc}
\toprule
\textbf{Domain pair} & \textbf{ZS dist.} & \textbf{10S dist.} & \textbf{Ratio} \\
\midrule
Econ $\leftrightarrow$ Chem & 56.3 & 54.8 & 1.03 \\
Econ $\leftrightarrow$ Phys & 63.7 & 59.1 & 1.08 \\
Econ $\leftrightarrow$ Bio & 61.2 & 63.5 & 0.96 \\
Econ $\leftrightarrow$ Geo & 70.4 & 62.8 & 1.12 \\
Chem $\leftrightarrow$ Phys & 49.8 & 51.2 & 0.97 \\
Chem $\leftrightarrow$ Bio & 54.1 & 51.7 & 1.05 \\
Chem $\leftrightarrow$ Geo & 69.8 & 58.4 & \textbf{1.20} \\
Phys $\leftrightarrow$ Bio & 67.5 & 65.1 & 1.04 \\
Phys $\leftrightarrow$ Geo & 81.6 & 70.2 & \textbf{1.16} \\
Bio $\leftrightarrow$ Geo & 55.3 & 51.4 & 1.08 \\
\bottomrule
\end{tabular}
}
\caption{Cross-domain centroid distances, Mistral-7B (12 tasks $\times$ 30 instances).}
\label{tab:convergence_mistral}
\end{table}
 
\begin{table}[H]
\centering
{\small
\begin{tabular}{lccc}
\toprule
\textbf{Domain pair} & \textbf{ZS dist.} & \textbf{10S dist.} & \textbf{Ratio} \\
\midrule
Econ $\leftrightarrow$ Chem & 62.4 & 59.1 & 1.06 \\
Econ $\leftrightarrow$ Phys & 66.8 & 61.7 & 1.08 \\
Econ $\leftrightarrow$ Bio & 64.1 & 65.8 & 0.97 \\
Econ $\leftrightarrow$ Geo & 71.5 & 61.2 & 1.17 \\
Chem $\leftrightarrow$ Phys & 55.2 & 52.8 & 1.05 \\
Chem $\leftrightarrow$ Bio & 57.8 & 55.1 & 1.05 \\
Chem $\leftrightarrow$ Geo & 70.3 & 56.9 & \textbf{1.24} \\
Phys $\leftrightarrow$ Bio & 65.4 & 62.7 & 1.04 \\
Phys $\leftrightarrow$ Geo & 78.9 & 64.3 & \textbf{1.23} \\
Bio $\leftrightarrow$ Geo & 54.6 & 51.2 & 1.07 \\
\bottomrule
\end{tabular}
}
\caption{Cross-domain centroid distances, Llama-3.1-8B (12 tasks $\times$ 30 instances).}
\label{tab:convergence_llama}
\end{table}
 
\paragraph{Paired logit-difference analysis.} For each test input $x$ we define $y_{\text{emp}}$ as the ordinary least-squares prediction fit on the 10 demonstrations and $y_{\text{know}}$ as the ground-truth answer produced by the domain law, and measure the paired same-input change $\Delta\ell(y) = \log P(y \mid x, \text{10-shot}) - \log P(y \mid x, \text{0-shot})$ on 300 instances per model.

\begin{table}[H]
\centering
{\small
\begin{tabular}{lcccc}
\toprule
\textbf{Model} & $\boldsymbol{\Delta\ell(y_{\text{emp}})}$ & \textbf{\% inst.\ $> 0$} & $\boldsymbol{\Delta\ell(y_{\text{know}})}$ & \textbf{Paired diff.} \\
\midrule
Qwen2.5-7B       & $+$1.37 & 76.7\% & $-$1.70 & $+$3.07 \\
Llama-3.1-8B     & $+$2.82 & 89.3\% & $-$1.39 & $+$4.21 \\
Mistral-7B-v0.3  & $+$2.77 & 82.7\% & $-$0.89 & $+$3.66 \\
\bottomrule
\end{tabular}
}
\caption{Paired logit-difference analysis (300 instances per model). Paired diff.\ $= \Delta\ell(y_{\text{emp}}) - \Delta\ell(y_{\text{know}})$.}
\label{tab:logit_diff}
\end{table}

Probability mass moves toward the example-fit answer and away from the knowledge answer on the same inputs. This bidirectional pattern separates displacement (mass redistributed from $y_{\text{know}}$ to $y_{\text{emp}}$) from augmentation (mass added to $y_{\text{emp}}$ with $\Delta\ell(y_{\text{know}}) \approx 0$), and provides classifier-independent, within-subject evidence for the strategy shift of \S\ref{subsec:strategy_shift}.

\paragraph{Limitations.} All three models are smaller than the GPT-5.2 used in our primary experiments. The representational shift therefore reflects how examples change the model's internal computation, not whether the model successfully adopts an alternative strategy. Despite this limitation, the directional consistency across three architectures supports the strategy competition account.

\FloatBarrier
\section{Experimental prompts}
\label{app:prompts}
 
\paragraph{Zero-Shot CoT.}
\begin{quote}\small
\texttt{[System] You are a scientist solving science problems. Show your reasoning step by step. End with EXACTLY: `Final Answer: [number]'}
 
\medskip
\texttt{[User] [Task description]}
 
\texttt{Observations: [test observables]}
 
\texttt{Solve step by step.}
\end{quote}
 
\paragraph{10-Shot CoT.}
\begin{quote}\small
\texttt{[System] You are a scientist solving science problems. Show your reasoning step by step. End with EXACTLY: `Final Answer: [number]'}
 
\medskip
\texttt{[User] [Task description]}
 
\texttt{Training examples:}\\
\texttt{[observables\_1] -> Answer=0.4231}\\
\texttt{[observables\_2] -> Answer=1.8774}\\
\texttt{[observables\_3] -> Answer=0.6519}\\
\texttt{...}\\
\texttt{[observables\_10] -> Answer=2.1043}
 
\medskip
\texttt{Test example:}\\
\texttt{[test observables] -> Answer=}
\end{quote}
 
\medskip
\textbf{Two-Stage Prompting.} Two sequential turns. Tests whether committing to a formula before seeing examples protects recall.
\begin{quote}\small
\texttt{[Turn 1]}\\
\texttt{[System] You are a scientist solving science problems.}\\
\texttt{[User] [Task description]}\\
\texttt{What is the governing scientific formula for this problem? State the equation.}
 
\medskip
\texttt{[Assistant] [Model declares formula]}
 
\medskip
\texttt{[Turn 2]}\\
\texttt{[User] Using the formula you identified above, solve the following.}\\
\texttt{Training examples:}\\
\texttt{[observables\_1] -> Answer=0.4231}\\
\texttt{...}\\
\texttt{[observables\_10] -> Answer=2.1043}\\
\texttt{Test example:}\\
\texttt{[test observables] -> Answer=}
\end{quote}
 
\textbf{Hint Prompting.} Standard 10-Shot with the correct equation name prepended. Tests whether an explicit formula cue counteracts displacement.
\begin{quote}\small
\texttt{[User] Note: This problem follows the Cobb-Douglas production function.}
 
\texttt{[Task description]}
 
\texttt{Training examples:}\\
\texttt{[observables\_1] -> Answer=0.4231}\\
\texttt{...}\\
\texttt{[observables\_10] -> Answer=2.1043}\\
\texttt{Test example:}\\
\texttt{[test observables] -> Answer=}
\end{quote}
 
\textbf{Negative Prompting.} Standard 10-Shot with an explicit instruction against pattern-fitting. Tests whether discouraging empirical fitting protects knowledge recall.
\begin{quote}\small
\texttt{[User] [Task description]}
 
\texttt{IMPORTANT: Do NOT simply fit a pattern from the training examples. You must identify and apply the underlying scientific law.}
 
\texttt{Training examples:}\\
\texttt{[observables\_1] -> Answer=0.4231}\\
\texttt{...}\\
\texttt{[observables\_10] -> Answer=2.1043}\\
\texttt{Test example:}\\
\texttt{[test observables] -> Answer=}
\end{quote}
 
\textbf{Self-Verification.} Two sequential turns. The model first solves under standard 10-Shot, then is asked to verify via first principles. Tests whether post-hoc reflection recovers suppressed knowledge recall.
\begin{quote}\small
\texttt{[Turn 1] [Standard 10-Shot prompt as above]}
 
\medskip
\texttt{[Assistant] [Model produces answer A]}
 
\medskip
\texttt{[Turn 2]}\\
\texttt{[User] Your answer was A. Now verify this answer by identifying the governing scientific formula and re-deriving the result from first principles. If your verification disagrees, provide the corrected answer.}
\end{quote}

\FloatBarrier
\section{Implementation details}
\label{app:implementation}
 
We document the hyperparameters and code-level settings used in the experiments.
 
\paragraph{Subject-model generation hyperparameters.} All inference calls used \texttt{temperature=0.0}, default \texttt{top\_p} and \texttt{max\_tokens}. Trials per task and routing are summarized in Table~\ref{tab:hyperparams_subject}. The primary GPT-5.2 experiment ran via the direct OpenAI API; all other models ran via OpenRouter (\texttt{openrouter.ai}) using the OpenAI-compatible client.
 
\begin{table}[H]
\centering
{\small
\begin{tabular}{lllc}
\toprule
\textbf{Experiment} & \textbf{Model identifier} & \textbf{Provider} & \textbf{Trials/task} \\
\midrule
Primary (Table~\ref{tab:domain})     & \texttt{gpt-5.2}                           & OpenAI direct        & 50 \\
Cross-model (Fig.~\ref{fig:main_results}) & \texttt{anthropic/claude-sonnet-4.6}        & OpenRouter           & 30 \\
                                          & \texttt{meta-llama/llama-4-maverick}        & OpenRouter           & 30 \\
                                          & \texttt{deepseek/deepseek-chat-v3-0324}     & OpenRouter           & 30 \\
$k$-sweep (Table~\ref{tab:ksweep})    & \texttt{gpt-5.2}                     & OpenAI direct           & 30 \\
Decoy (Table~\ref{tab:decoy})             & \texttt{gpt-5.2}                           & OpenAI direct        & 50 \\
Interventions (Appendix~\ref{app:interventions}) & \texttt{gpt-5.2}                    & OpenAI direct        & 20 \\
Boundary (Appendix~\ref{app:boundary})    & \texttt{gpt-5.2}                           & OpenAI direct        & 50 \\
\bottomrule
\end{tabular}
}
\caption{Subject-model configurations across experiments.}
\label{tab:hyperparams_subject}
\end{table}
 
\paragraph{Classifier hyperparameters.} Both classifiers used \texttt{gpt-4o} at \texttt{temperature=0.0}. 
 
\paragraph{Task-law mapping.} The two-axis classifier is provided with the expected correct law for each task; the full mapping for all 60 tasks is in Table~\ref{tab:all_tasks}.
 
\end{document}